\documentclass{article}

\usepackage{microtype}
\usepackage{graphicx}
\usepackage{subfigure}
\usepackage{algpseudocode}
\usepackage{booktabs} 
\usepackage{natbib}
\usepackage{times}
\usepackage{epsfig}
\usepackage{graphicx}
\usepackage{dsfont}
\usepackage{amsmath}
\usepackage{amssymb}
\graphicspath{{./images/}}
\usepackage{tabularx}
\usepackage{makecell}
\usepackage{bm}
\usepackage{multirow} 
\usepackage{rotating}
\usepackage{wrapfig}
\usepackage{subfigure}
\usepackage{amsthm}
\usepackage{thmtools}
\usepackage{thm-restate}
\usepackage{flexisym}
\usepackage{colortbl}
\usepackage{xcolor}
\usepackage{appendix}

\newcommand{\BU}[1]{\color{black!40!green}\textbf{#1}}

\newcommand{\BV}[1]{\color{black!10!blue}{\em #1}}

\usepackage{hyperref}

\usepackage[accepted]{icml2019}

\begin{document}

\twocolumn[
\icmltitle{Domain Agnostic Learning with Disentangled Representations}

\begin{icmlauthorlist}
\icmlauthor{Xingchao Peng}{bu}
\icmlauthor{Zijun Huang}{col}
\icmlauthor{Ximeng Sun}{bu}
\icmlauthor{Kate Saenko}{bu}
\end{icmlauthorlist}

\icmlaffiliation{bu}{Computer Science Department, Boston University; 111 Cummington Mall, Boston, MA 02215, USA; email:xpeng@bu.edu}
\icmlaffiliation{col}{Columbia Unversity and MADO AI Research; 116th St and Broadway, New York, NY 10027, USA; email:zijun.huang@columbia.edu}
\icmlcorrespondingauthor{Kate Saenko}{saenko@bu.edu}

\icmlkeywords{domain-agnostic learning, domain adaptation, feature disentanglement}

\vskip 0.3in
]

\printAffiliationsAndNotice{}  

\begin{abstract}
Unsupervised model transfer  has the potential to greatly improve the generalizability of deep models to novel domains. Yet the current literature assumes that the separation of target data into distinct domains is known as a priori. In this paper, we propose the task of Domain-Agnostic Learning (DAL): How to  transfer knowledge from a labeled source domain to unlabeled data from arbitrary target domains? To tackle this problem, we devise a novel Deep Adversarial Disentangled Autoencoder (DADA) capable of disentangling domain-specific features from class identity. We demonstrate experimentally that when the target domain labels are unknown, DADA leads to state-of-the-art performance on several image classification datasets.
\end{abstract}

\section{Introduction}
\label{sec_intro}

Supervised machine learning assumes that training and testing data are sampled \textit{i.i.d} from the same distribution, while in practice, the training and testing data are typically collected from related domains but under different distributions, a phenomenon known as \textit{domain shift}~\cite{datashift_book2009}. To avoid the cost of annotating each new test domain, \textit{Unsupervised Domain Adaptation (UDA)} tackles domain shift by aligning the feature distribution of the source domain with that of the target domain, resulting in domain-invariant features. 
However, current methods assume that target samples have domain labels and therefore can be isolated into separate homogeneous domains. For many practical applications, this is an overly strong assumption. For example, a hand-written character recognition system could encounter characters written by different people, on different materials, and under different lighting conditions; an image recognition system applied to images scraped from the web must handle mixed-domain data (\textit{e.g.} paintings, sketches, clipart) without 
their domain labels.

In this paper, we consider Domain-Agnostic Learning (DAL), a more difficult but practical problem of knowledge transfer from one labeled source domain to multiple unlabeled target domains. The main challenges of domain-agnostic learning are that: (1) the target data has mixed domains, which hampers the effectiveness of mainstream feature alignment methods~\cite{long2015, SunS16a, MCD_2018}, and 
(2) class-irrelevant information leads to \textit{negative transfer}~\cite{pan2010survey}, especially when the target domain is highly heterogeneous. 

Mainstream UDA methods align the source domain to the target domain by minimizing the Maximum Mean Discrepancy~\cite{long2015,ddc}, aligning high-order moments~\cite{SunS16a, cmd}, or adversarial training~\cite{DANN, adda}. However, these methods are designed for one-to-one domain alignment and do not account for multiple latent domains in the target. \textit{Multi-source domain adaptation}~\cite{domainnet, xu2018deep, Mansour_nips2018} considers adaptation between multiple sources and a single target domain and assumes domain labels on the source data. 
\textit{Continuous domain adaptation}~\cite{continuous_DA} aims to transfer knowledge to a continuously changing domain (\textit{e.g.} cars in different decades), but in their scenario the target data are temporally related.
Recently, \textit{domain generalization} approaches~\cite{li2018domain,domain_generalization, li2018domain_} have been introduced to adapt from multiple labeled source domains to an unseen target domain. All of the above models make a strong assumption that the target data are homogeneously sampled from the same distribution, unlike the scenario we consider here. 

We postulate that a solution to domain-agnostic learning should not only learn invariance between source and target, but should also actively disentangle the class-specific features from the remaining information  in the image.
Deep neural networks are known to extract features in which multiple hidden factors are highly entangled~\cite{bengio2013representation}. 
Recent work attempts to disentangle features in the latent space of autoencoders with adversarial training~\cite{dida, detachandattach, cisac_gan, drit}. However, the above models have limited capacity in transferring features learned from one domain to heterogeneous target domains. 
~\citet{ufdn} proposes a framework that takes samples from multiple domains as input, and derives a domain-invariant latent feature space via adversarial training. This model is limited by two factors when applied to the DAL task. First, it only disentangles the embeddings into domain-invariant features and domain-specific features such as weather conditions, and discards the latter, 
but does not explicitly try to separate class-relevant features from class-irrelevant features like background.
Second, there is no guarantee that the domain-invariant features are fully disentangled from the domain-specific features.

\begin{figure*}[t]
    \centering
    \includegraphics[width=.9\linewidth]{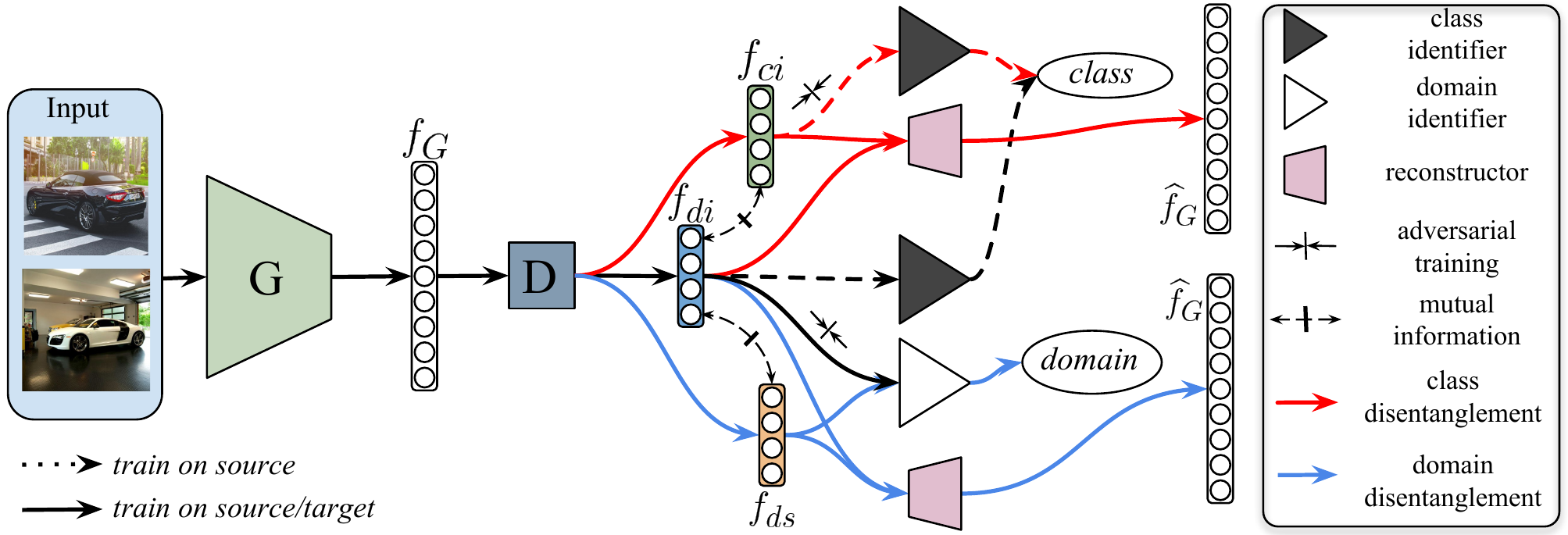}
    \vspace{-0.3cm}
    \caption{Our DADA architecture learns to extract \textit{domain-invariant} features of visual categories. In addition to \textit{domain disentanglement} ({\color{blue}blue lines}), we employ \textit{class disentanglement} ({\color{red}red lines}) to remove  \textit{class-irrelevant} features, both trained adversarially. 
    We further apply a mutual information minimizer to strengthen the disentanglement. }
   \vspace{-0.3cm}
    \label{fig_overview}
\end{figure*}

To address the issues mentioned above, we propose a novel \textit{Deep Adversarial Disentangled Autoencoder (DADA)}, aiming to tackle domain-agnostic learning by disentangling the domain-invariant features from both domain-specific and class-irrelevant features simultaneously. First, in addition to \textit{domain disentanglement}~\cite{ufdn,dida,drit}, we employ \textit{class disentanglement} to remove class-irrelevant features, as shown in Figure~\ref{fig_overview}. The \textit{class disentanglement} is trained in an adversarial fashion: a class identifier is trained on the labeled source domain and the disentangler generates features to fool the class identifier. To the best of our knowledge, we are the first to show that \textit{class disentanglement} boosts domain adaptation performance. Second, to enhance the disentanglement, we propose to minimize the mutual information between the disentangled features. We implement a neural network to estimate the mutual information between the disentangled feature distributions, inspired by a recently published theoretical work~\cite{mine}.
Comprehensive experiments on standard image recognition datasets demonstrate that our derived disentangled representation achieves significant improvements over the state-of-the-art methods on the task of domain-agnostic learning.

The main contributions of this paper are highlighted as follows: (1) we propose a novel learning paradigm of domain-agnostic learning; 2) we develop an end-to-end Deep Adversarial Disentangled Autoencoder (DADA) which learns a better disentangled feature representation to tackle the task; and (3) We propose \textit{class disentanglement} to remove class-irrelevant features, and minimize the mutual information to enhance the disentanglement.

\section{Related Work}

\textbf{Domain Adaptation} Unsupervised domain adaptation (UDA) aims to transfer the knowledge learned from one or more labeled source domains to an unlabeled target domain. Various methods have been proposed, including discrepancy-based UDA approaches~\cite{JAN, ddc, ghifary2014domain, peng2017synthetic}, adversary-based approaches~\cite{cogan, adda, ufdn}, and reconstruction-based approaches~\cite{yi2017dualgan, CycleGAN2017, hoffman2017cycada, kim2017learning}. These models are typically designed to tackle single source to single target adaptation. Compared with single source  adaptation, multi-source domain adaptation (MSDA) assumes that training data are collected from multiple sources. Originating from the theoretical analysis in~\cite{ben2010theory, Mansour_nips2018, crammer2008learning}, MSDA has been applied to many practical applications~\cite{xu2018deep, duan2012exploiting, domainnet}. Specifically, ~\citet{ben2010theory} introduce an $\mathcal{H}\Delta\mathcal{H}$-divergence between the weighted combination of source domains and a target domain. 
We propose a new and more practical learning paradigm, not yet considered in the UDA literature, where labeled data come from a single source domain but the testing data contain multiple unknown domains. 

\textbf{Representation Disentanglement} The goal of learning disentangled representations is to model the factors of data variation. Recent works~\cite{mathieu2016disentangling, makhzani2015adversarial, ufdn, cisac_gan} aim at learning an interpretable representation using generative adversarial networks (GANs)~\cite{gan, kingma2014semi} and variational  autoencoders (VAEs)~\cite{stochastic_icml_2014, vae}. In a fully supervised setting, \citet{drit} proposes to disentangle the feature representation into a domain-invariant content space and a domain-specific attribute space, producing diverse outputs without paired training images. Another work~\cite{cisac_gan} proposes an auxiliary classifier GAN (AC-GAN) to achieve representation disentanglement. Despite promising performance, these methods focus on disentangling representation in a single domain. ~\citet{ufdn} introduces a unified feature disentangler to learn a domain-invariant representation from data across multiple domains. However, their model assumes that multiple source domains are available during training, which limits its practical application. In contrast, our model disentangles the representation based on one source domain and multiple unknown target domains, and proposes an improved approach to disentanglement that considers the class label and mutual information between features.

\textbf{Agnostic Learning} There are several prior studies of agnostic learning that are 
related to our work. Model-Agnostic Meta-Learning (MAML) \cite{maml} aims to train a model on a variety of learning tasks and solve a new task using only a few training examples. Different from MAML, our method mainly focuses on transferring knowledge to heterogeneous domains. \citet{agnostic_dg_2018} proposes a learning framework to seamlessly extend the knowledge from multiple source domain to an unseen target domain by pixel-adaptation in an incremental architecture. \citet{domain_agnostic_norm18} introduces a domain agnostic normalization layer for  adversarial UDA and improves the performance of deep models on an unseen domain. Though the results are promising, we argue that only normalizing the feature representation is not enough for domain-agnostic learning, and that extracting disentangled domain-invariant and domain-specific features is also important. 
\section{DADA: Deep Adversarial Disentangled Autoencoder}
\label{dada}

We define the domain-agnostic learning task as follows: Given a \emph{source} domain $\widehat{\mathcal{D}}_s = \{(\mathbf{x}_i^s,y^s_i)\}_{i=1}^{n_s}$ with $n_s$ labeled examples, the goal is to minimize risk on $N$ \emph{target} domains $\widehat{\mathcal{D}}_t$ = \{$\widehat{\mathcal{D}}_1, \widehat{\mathcal{D}}_2, ... ,\widehat{\mathcal{D}}_N\}$ without domain labels. We denote the target domains as $\widehat{\mathcal{D}}_t = \{\mathbf{x}_j^t\}_{j=1}^{n_t}$ with $n_t$ unlabeled examples. 
Empirically, we want to minimize the target risk ${\epsilon_t}\left( \theta  \right) = {\Pr _{\left( {{\mathbf{x}},y} \right) \sim \widehat{\mathcal{D}}_t }}\left[ {\theta \left( {\mathbf{x}} \right) \ne y} \right]$, where $\theta\left(\mathbf{x}\right)$ is the classifier.

We propose to solve the task by learning \textit{domain-invariant} features that are discriminative of the class.
Figure~\ref{fig_overview} shows the proposed model. The feature generator $G$ maps the input image to a feature vector $f_G$, which has many highly entangled factors. The disentangler $D$ is responsible for disentangling the features ($f_G$) 
into \textit{domain-invariant} features ($f_{di}$), \textit{domain-specific} features ($f_{ds}$), and \textit{class-irrelevant} features ($f_{ci}$). The feature reconstructor $R$ aims to recover $f_G$ from either ($f_{di}$, $f_{ds}$) or ($f_{di}$, $f_{ci}$).
\textit{D} and \textit{R} are implemented as the encoder and decoder in a Variational Autoencoder. A mutual information minimizer is applied between $f_{di}$ and $f_{ci}$, as well as between $f_{di}$ and $f_{ds}$, to enhance the disentanglement. Adversarial training via a domain identifier aligns the source domain and the heterogeneous target domain in the $f_{di}$ space. A class identifier $C$ is trained on the labeled source domain 
to predict the class distribution $f_C$ and to adversarially extract class-irrelevant features $f_{ci}$. We next describe each component in detail.

\textbf{Variational Autoencoders} VAEs~\cite{vae} are a class of deep generative models that simultaneously train both a probabilistic encoder and decoder. The encoder is trained to generate latent vectors that roughly follow a \textit{Gaussian} distribution. 
In our case, we learn each part of our disentangled representations by applying a VAE architecture with the following objective function:
\begin{equation} 
\label{eqn:vae}
\mathcal{L}_{vae} = {\lVert \widehat{f}_G - f_G \rVert}_{F}^{2} +KL(q(z|f_G)||p(z)), 
\end{equation}
where the first term aims at recovering the original features extracted by $G$, and the second term calculates \textit{Kullback-Leibler divergence} which penalizes the deviation of latent features from the prior distribution $p(z_c)$ (as $z \sim \mathcal{N}(0,I)$). However, this property cannot guarantee that \textit{domain-invariant} features are well disentangled from the \textit{domain-specific} features or from \textit{class-irrelevant} features, as the loss function in Equation~\ref{eqn:vae} only aligns the latent features to a normal distribution.

\textbf{Class Disentanglement} To address the above problem, we employ \textit{class disentanglement} to remove class-irrelevant features, such as background, in an adversarial way.
First, we train the disentangler $D$ and the $K$-way class identifier $C$ to correctly predict the labels, supervised by the cross-entropy loss:
\begin{equation}
    \mathcal{L}_{ce} = -\mathbb{E}_{(x_s,y_s)\sim\widehat{\mathcal{D}}_s} \sum_{k=1}^{K}\mathds{1} [k=y_s]log(C(f_D))
    \label{equ_cross_entropy}
\end{equation}
where $f_D\in\{f_{di}, f_{ci}\}$.

In the second step, we fix the class identifier and train the disentangler $D$ to fool the class identifier by generating class-irrelevant features $f_{ci}$. This can be achieved by minimizing the negative entropy of the 
predicted class distribution:
\begin{equation}
\mathcal{L}_{ent} = - \frac{1}{n_s} \sum_{j=1}^{n_s} \log C(f^j_{ci}) - \frac{1}{n_t} \sum_{j=1}^{n_t} \log C(f^j_{ci})
\label{equ_entropy_loss}
\end{equation}
where the first term and the second term indicate minimizing the entropy on the source domain and on heterogeneous target, respectively. 
The above adversarial training process forces the corresponding \textit{disentangler} to extract \textit{class-irrelevant} features.

\textbf{Domain Disentanglement} To tackle the domain agnostic learning task, 
disentangling class-irrelevant features is not enough, as it fails to align the source domain with the target. To achieve better alignment, we further propose 
to disentangle the learned features into \textit{domain-specific} and \textit{domain-invariant} and to thus align the source with the target domain in the \textit{domain-invariant} latent space. This is achieved by exploiting adversarial domain classification in the resulting latent space. Specifically, we leverage a domain identifier $DI$, which takes the disentangled feature ($f_{di}$ or $f_{ds}$ ) as input and outputs the domain label $l_f$ (source or target). The objective function of the domain identifier is as follows:
\begin{equation}
\label{eqn_domain_identifier}
\mathcal{L}_{DI} = - \mathbb{E}[l_f\log P( l_f)]  + \mathbb{E}(1-l_f)[\log P(1-l_f )],
\end{equation}
Then the disentangler is trained to fool the domain identifier $DI$ to extract domain-invariant features.

\begin{figure*}[t]
    \centering
    \includegraphics[width=\linewidth]{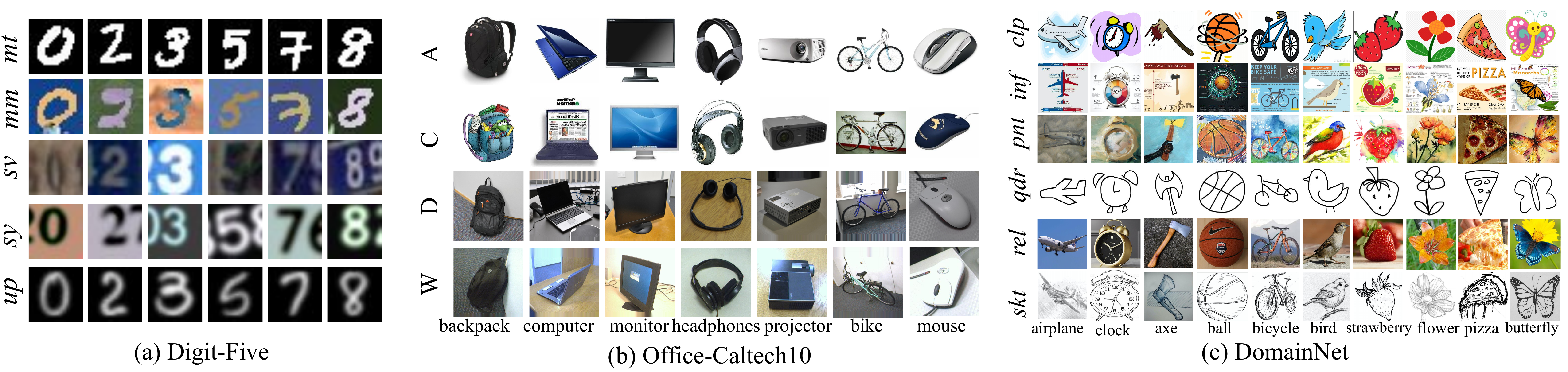}
    \vspace{-0.6cm}
    \caption{We demonstrate the effectiveness of DADA on three dataset: Digit-Five, Office-Caltech10~\cite{gong2012geodesic} and DomainNet~\cite{domainnet} dataset. The Digit-Five dataset includes: MNIST (\textit{mt}), MNIST-M (\textit{mm}), SVHN (\textit{sv}), Synthetic (\textit{syn}), and USPS (\textit{up}). The Office-Caltech10 dataset contains: \textit{Amazon} (A), \textit{Caltech} (C), \textit{DSLR} (D), and \textit{Webcam} (W). The DomainNet dataset includes: \textit{clipart} (\textit{clp}), \textit{infograph} (\textit{inf}), \textit{painting} (\textit{pnt}), \textit{quickdraw} (\textit{qdr}), \textit{real} (\textit{rel}), and \textit{sktech} (\textit{skt}).} 

    \label{fig_dataset_overview}
\end{figure*}
\newcommand{\PJ}[2]{\mathbb{P}_{#1#2}} 
\newcommand{\PJE}[2]{\mathbb{P}^{n}_{#1#2}}  
\newcommand{\PI}[2]{\mathbb{P}_{#1}\otimes\mathbb{P}_{#2}}  
\textbf{Mutual Information Minimization} To better disentangle the features, we minimize the mutual information between \textit{domain-invariant} and \textit{domain-specific} features ($f_{di}$, $f_{ds}$), as well as \textit{domain-invariant} and \textit{class-irrelevant} features ($f_{di}$, $f_{ci}$):
\begin{align}
I(\mathcal{D}_{x}; \mathcal{D}_{f_{di}}) = \int_{\mathbb{X} \times \mathcal{Z}} \log{\frac{d\PJ{X}{Z}}{d\PI{X}{Z}}} d\PJ{X}{Z},
\label{eq:mutual_information}
\end{align}

where $x\in\{f_{ds},f_{ci}\}$, $\PJ{X}{Z}$ is the joint probability distribution of ($\mathcal{D}_{x}$, $\mathcal{D}_{f_{di}}$), and $\mathbb{P}_X =
\int_{\mathcal{Z}} d\PJ{X}{Z}$ and $\mathbb{P}_Z = \int_{\mathcal{X}} d\PJ{X}{Z}$ are the marginals. Despite being a pivotal measure across different domains, the mutual information is only tractable for discrete variables, or for a limited family of problems where the probability distributions are unknown~\cite{mine}. The computation incurs a complexity of $O(n^2)$, which is undesirable for deep CNNs. Is this paper, we adopt the Mutual Information Neural Estimator (MINE)~\cite{mine}

\begin{equation}
    \label{donskeremp}
    \widehat{I(X;Z)}_n = 
    \sup_{\theta \in \Theta} \mathbb{E}_{\mathbb{P}^{(n)}_{XZ}}[T_\theta] - \log(\mathbb{E}_{\mathbb{P}^{(n)}_{X} \otimes \widehat{\mathbb{P}}^{(n)}_{Z}}[e^{T_\theta}]).
\end{equation} 

which provides unbiased estimation of mutual information on $n$ \textit{i.i.d} samples by leveraging a neural network $T_\theta$.

\begin{algorithm}[t]			
	\caption{Learning algorithm for DADA} \label{alg_DADA}
	\begin{small}
		\hspace*{0.02in}{\bf Input:} source labeled datasets $\{(\mathbf{x}_i^s,y^s_i)\}_{i=1}^{n_s}$; heterogeneous target dataset $\{\mathbf{x}_j^t\}_{j=1}^{n_t}$; feature extractor $G$; disentangler $D$; category identifier $C$, domain identifier $DI$, mutual information estimator $M$, and reconstructor $R$.\\
		\hspace*{0.02in}{\bf Output:} well-trained feature extractor $\hat{G}$, well-trained disentangler $\hat{D}$, and class identifier $\hat{C}$.
		\begin{algorithmic}[1]
			\While{not converged}
			\State Sample mini-batch from $\{(\mathbf{x}_i^s,y^s_i)\}_{i=1}^{n_s}$ and $\{\mathbf{x}_j^t\}_{j=1}^{n_t}$;
			\State \textbf{Class Disentanglement:}
			\For{1:iter}
			
			\State Update $G$, $D$, $C$ by Eq.\ref{equ_cross_entropy};
			\State Update $D$ by Eq.\ref{equ_entropy_loss};
			\EndFor
			\State \textbf{Domain Disentanglement:}
			\State Update $D$ and $DI$ by Eq.\ref{eqn_domain_identifier};
			\State \textbf{Mutual Information Minimization}:
			\State Calculate mutual information between the disentangled feature pair ($f_{di}$, $f_{ds}$), as well as ($f_{di}$,$f_{ci}$) with $M$;
			\State Update $D$, $M$ by Eq.\ref{equ_mutual_information};
			\State \textbf{Reconstruction:}
			\State Reconstruct $f_G$ by ($f_{di}$,$f_{ci}$) and ($f_{di}$, $f_{ds}$) with $R$;
			\State Update $D$, $R$ by Eq.\ref{eqn:vae}
			\EndWhile \\
			\Return $\hat{G} = G; \hat{C} = C; \hat{D} = D$.
		\end{algorithmic}
	\end{small}
\end{algorithm}

Practically, MINE (\ref{donskeremp}) can be computed as $I(X;Z)=\int\int{\mathbb{P}^{n}_{XZ}(x,z)T(x,z,\theta)}$ - $\log (\int\int\mathbb{P}^{n}_{X}(x)\mathbb{P}^n_{Z}(z)e^{T(x,z,\theta)})$. Additionally, to avoid computing the integrals, we leverage Monte-Carlo integration:
\begin{equation}
I(X,Z)=\frac{1}{n}\sum^{n}_{i=1}T(x,z,\theta)-\log(\frac{1}{n}\sum_{i=1}^{n}e^{T(x,z',\theta)})
\label{equ_mutual_information}
\end{equation}
 where $(x,z)$ are sampled from the joint distribution and $z'$ is sampled from the marginal distribution. We implement a neural network to perform the Monte-Carlo integration defined in Equation~\ref{equ_mutual_information}.

\textbf{Ring-style Normalization} Conventional batch normalization~\cite{batch_normalization} diminishes internal covariate shift by subtracting the batch mean and dividing by the batch standard deviation. Despite promising results on domain adaptation, batch normalization alone is not enough to guarantee that the embedded features are well normalized in the scenario of heterogeneous domains. The target data are sampled from multiple domains and their embedded features are scattered irregularly in the latent space. \citet{ringloss} proposes a ring-style norm constraint to maintain a balance between the angular classification margins of multiple classes. Its objective is as follows:
\begin{equation}
    \mathcal{L}_{ring} =  \frac{1}{2n}\sum_{i=1}^{n}(||T(x_i)||_2-R)^2
\end{equation}
where $R$ is the learned norm value. However, ring loss is not robust and may cause mode collapse if the learned $R$ is small. Instead, we incorporate the ring loss into a Geman-McClure model and minimize the following loss function:
\begin{equation}
    \mathcal{L}_{ring}^{GM} =  \frac{\sum_{i=1}^{n}(||T(x_i)||_2-R)^2}{2n\beta +\sum_{i=1}^{n}(||T(x_i)||_2-R)^2 }
    \label{equ_ring_loss}
\end{equation}
where $\beta$ is the scale factor of the Geman-McClure model. 

\textbf{Optimization} Our model is trained in an end-to-end fashion. We train the \textit{class} and \textit{domain disentanglement} component, MINE and the reconstruction component iteratively with Stochasitc Gradient Descent~\cite{SGD} or Adam~\cite{Adam} optimizer. We employ the popular neural networks (\textit{e.g.} LeNet, AlexNet, or ResNet) as our feature generator $G$. The detailed training procedure is presented in Algorithm~\ref{alg_DADA}.

\section{Experiments}

\newcommand{\RomanNumeralCaps}[1]
    {\MakeUppercase{\romannumeral #1}}

\begin{table*}
\caption{Accuracy on ``Digit-Five'' dataset with domain agnostic learning protocol. DADA achieves \textbf{62.3}\% accuracy, significantly outperforming other baselines. We incrementally add each component to our model, aiming to study their effectiveness on the final results. (model \textbf{\RomanNumeralCaps{1}}: with \textit{class disentanglement}; model \textbf{\RomanNumeralCaps{2}}: \textbf{\RomanNumeralCaps{1}} + \textit{domain disentanglement}; model \textbf{\RomanNumeralCaps{3}}: \textbf{\RomanNumeralCaps{2}} + ring loss; model \textbf{\RomanNumeralCaps{4}}: \textbf{\RomanNumeralCaps{3}} + reconstruction loss. {\BV{mt}}, {\BV{up}}, {\BV{sv}}, {\BV{sy}}, {\BV{mm}} are abbreviations for \textit{MNIST}, \textit{USPS}, \textit{SVHN}, \textit{Synthetic Digits}, \textit{MNIST-M}.) } \label{table_digit_five}
\vspace{0.1in}
\centering
{
\begin{tabular}{c c c c c c  c}
\Xhline{1pt} 
{Models} &
 {\scriptsize{\BV{mt$\rightarrow$mm,sv,sy,up}} } &  
 {\scriptsize{\BV{mm$\rightarrow$mt,sv,sy,up}} } & 
 {\scriptsize{\BV{sv$\rightarrow$mt,mm,sy,up}} }& 
 {\scriptsize{\BV{sy$\rightarrow$mt,mm,sv,up}} }&    
 {\scriptsize{\BV{up$\rightarrow$mt,mm,sv,sy}} } & 
{{\BU{Avg}}} \\ 

\hline

Source Only &20.5$\pm$1.2 &53.5$\pm$0.9 &62.9$\pm$0.3 &77.9$\pm$0.4 &22.6$\pm$0.4 & 47.5 \\
DAN~\cite{long2015} & 21.7$\pm$1.0 & 55.3$\pm$0.7 & 63.2$\pm$0.5 & 79.3$\pm$0.2 & 40.2$\pm$0.4 & 51.9 \\
DANN~\cite{DANN} & 22.8$\pm$1.1 & 45.2$\pm$0.6 & 61.8$\pm$0.2 & 79.3$\pm$0.3 & 38.7$\pm$0.6 & 49.6 \\
 ADDA~\cite{adda} & 23.4$\pm$1.3 & 54.8$\pm$0.8 & 63.5$\pm$0.4 & 79.6$\pm$0.3 & 43.5$\pm$0.5& 52.9  \\

UFDN~\cite{ufdn}& 20.2$\pm$1.5 & 41.6$\pm$0.7 & 64.5$\pm$0.4 & 60.7$\pm$0.3 & 44.6$\pm$0.2 & 46.3 \\
 MCD~\cite{MCD_2018}&28.7$\pm$1.3 &43.8$\pm$0.8 &75.1$\pm$0.3 &78.9$\pm$0.3 &\textbf{55.3}$\pm$0.4 & 56.4 \\

\Xhline{0.7pt} 

DADA+\textit{class} (\textbf{\RomanNumeralCaps{1}})  & 28.9$\pm$1.2 & 50.1$\pm$0.9 & 65.4$\pm$0.2 & 79.8$\pm$0.1 & \underline{50.4}$\pm$0.3 & 54.9 \\
 DADA+\textit{domain} (\RomanNumeralCaps{2})  & 34.1$\pm$1.7 & 57.1$\pm$0.4 & 71.3$\pm$0.4 & 82.5$\pm$0.3 & 45.4$\pm$0.4 & 57.5 \\ 
 DADA+\textit{ring} (\RomanNumeralCaps{3})  & \underline{35.3}$\pm$1.5 & \underline{57.5}$\pm$0.6 & \underline{80.1}$\pm$0.3 & \underline{82.9}$\pm$0.2 & 46.2$\pm$0.3 & 60.4 \\ 
DADA+\textit{rec} (\RomanNumeralCaps{4})  & \textbf{39.4}$\pm$1.4 & \textbf{61.1}$\pm$0.7 & \textbf{80.1}$\pm$0.4 & \textbf{83.7}$\pm$0.2 & 47.2$\pm$0.4 & \textbf{62.3}\\ 
\Xhline{1.0pt}
\end{tabular}
} 
\vspace{-0.4cm}

\end{table*}

\begin{figure*}[t]
    \begin{minipage}{\hsize}
      \centering
      \subfigure[\scriptsize Source Features ]
      {\includegraphics[width=0.23\hsize]{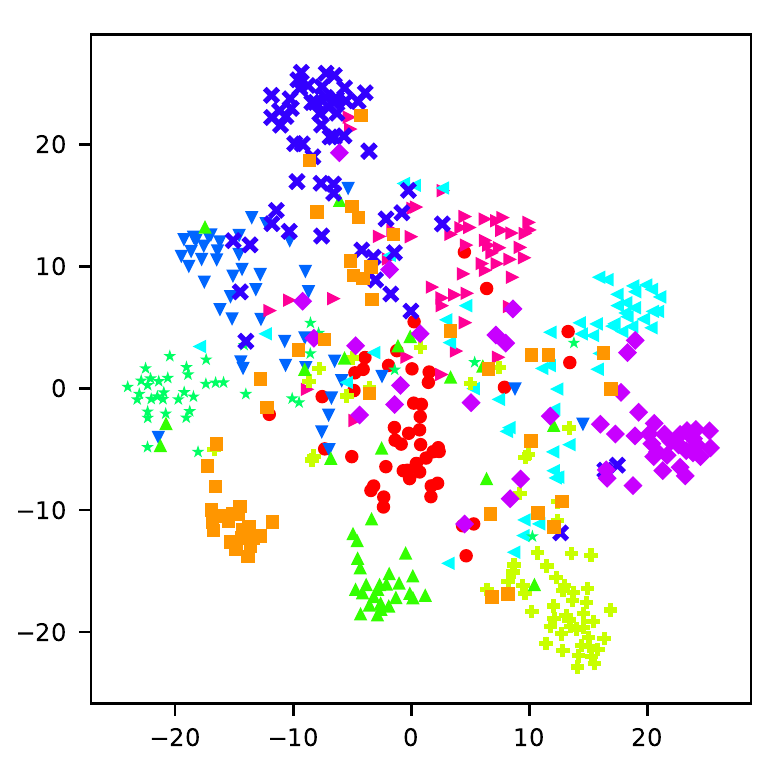}
      \label{tsne_source} }
     \centering
      \subfigure[\scriptsize UFDN Features]
      {\includegraphics[width=0.232\hsize]{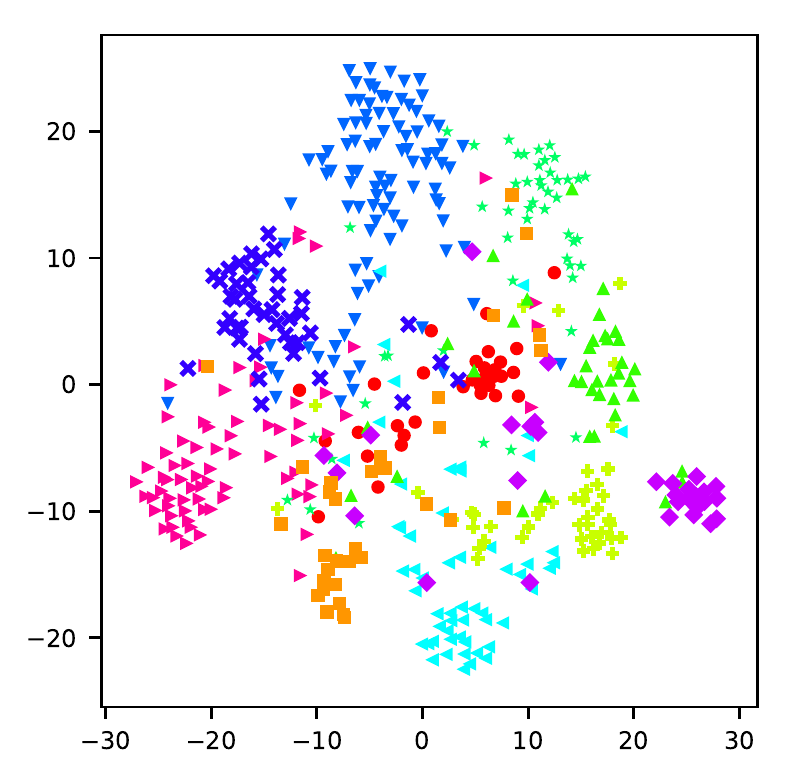}
      \label{tsne_ufdn}}
      \centering
      \subfigure[\scriptsize MCD Features ]
      {\includegraphics[width=0.228\hsize]{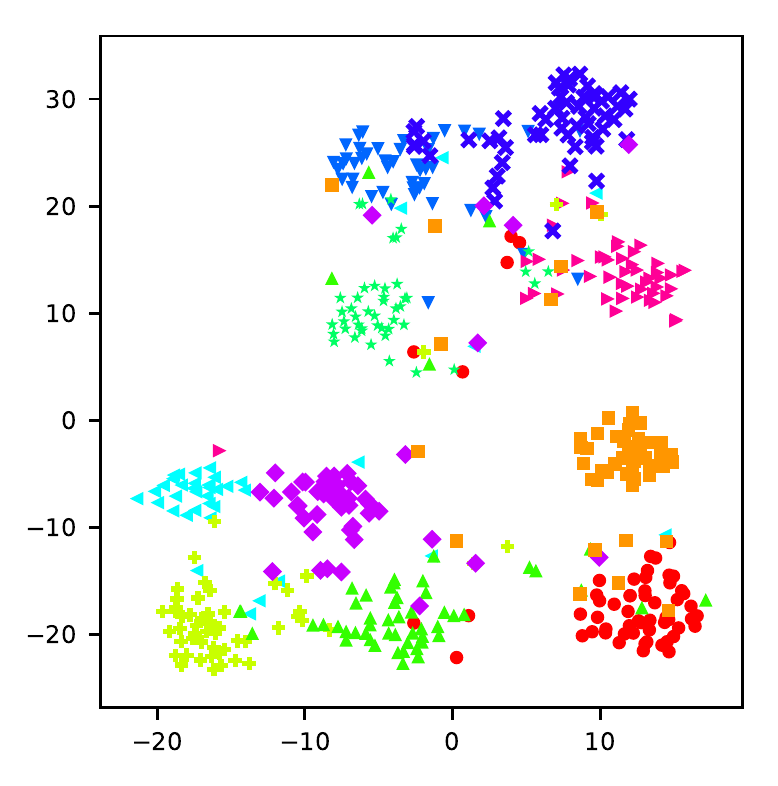} 
      \label{tsne_mcd}}
      \centering
      \subfigure[\scriptsize DADA Features ]
      { \includegraphics[width=0.232\hsize]{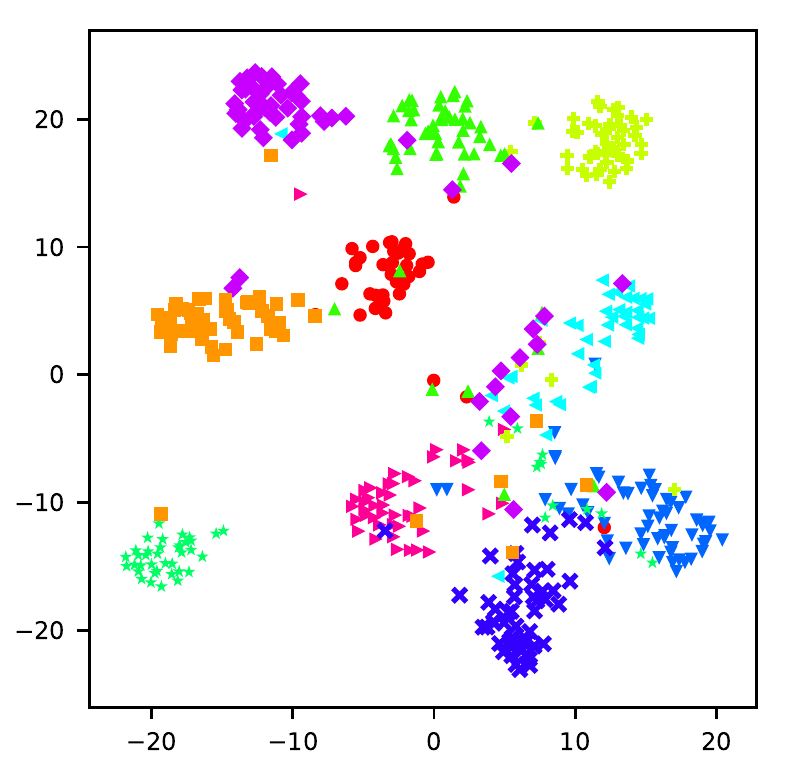}
      \label{tsne_dada}}
    \end{minipage}
    \vspace{-0.4cm}
  \caption{Feature visualization: t-SNE plot of source features, UFDN~\cite{ufdn} features, MCD~\cite{MCD_2018} features and DADA features on agnostic target domain in {\BV{sv}} $\rightarrow${\BV{mm},\BV{mt},\BV{up},\BV{sy}} setting. We use different markers and different colors to denote different categories. (Best viewed in color.)}
  \label{fig_analysis}
  
\end{figure*}

We compare the DADA model to state-of-the-art domain adaptation algorithms on the following tasks: digit classification (\textit{MNIST, SVHN, USPS, MNIST-M, Synthetic Digits}) and image recognition (Office-Caltech10~\cite{gong2012geodesic}, DomainNet~\cite{domainnet}). Sample images of these datasets can be seen in Figure~\ref{fig_dataset_overview}. Table~\ref{tab:dataset_details} (suppementary material) shows the detailed number of images we use in our experiments. In the main paper, we only report major results; more implementation details are provided in the supplementary material. All of our experiments are implemented in the PyTorch\footnote{\url{http://pytorch.org}} platform. 

\vspace{-0.3cm}
\subsection{Experiments on Digit Recognition}
\label{exp_digit}
\textbf{Digit-Five} This dataset is a collection of five benchmarks for digit recognition, namely \textit{MNIST}~\cite{lecun1998gradient}, \textit{Synthetic Digits}~\cite{DANN}, \textit{MNIST-M}~\cite{DANN}, \textit{SVHN}, and \textit{USPS}. In our experiments, we take turns setting one domain as the source domain and the rest as the mixed target domain (discarding both the class and the domain labels), leading to five transfer tasks. To explore the effectiveness of each component in our model, we propose four different ablations, \textit{i.e.} model \textbf{\RomanNumeralCaps{1}}: with \textit{class disentanglement}; model \textbf{\RomanNumeralCaps{2}}: \textbf{\RomanNumeralCaps{1}} + \textit{domain disentanglement}; model \textbf{\RomanNumeralCaps{3}}: \textbf{\RomanNumeralCaps{2}} + ring loss; model \textbf{\RomanNumeralCaps{4}}: \textbf{\RomanNumeralCaps{3}} + reconstruction loss. The detailed architecture of our model can be seen in Table~\ref{tab:digit_arch} (supplementary material).

\begin{table*}[t]

    \addtolength{\tabcolsep}{1pt}
    \caption{Accuracy on \emph{Office-Caltech10} dataset with DAL protocal. The methods in the above table are based on ``AlexNet'' backbone and the methods below are based on the ``ResNet'' backbone. For both backbones, our model outperforms other baselines. } \label{table_office}
    \vspace{0.1in}
    \centering
    
    \label{table:office10}
    \begin{tabular}{cccccc}
        \Xhline{1pt}
        Method & A $\rightarrow$ C,D,W & C $\rightarrow$ A,D,W & D $\rightarrow$ A,C,W & W $\rightarrow$ A,C,D & Average \\
        \hline
        AlexNet~\cite{alexnet}& 83.1$\pm$0.2 & 88.9$\pm$0.4 & 86.7$\pm$0.4  & 82.2$\pm$0.3& 85.2 \\
        DAN~\cite{long2015} & 82.5$\pm$0.3 & 86.2$\pm$0.4& 75.7$\pm$0.5& 80.4$\pm$0.2 & 81.2 \\
        
        RTN~\cite{RTN} & 85.2$\pm$0.4 & 89.8$\pm$0.3 & 81.7$\pm$0.3& 83.7$\pm$0.4 & 85.1 \\
        JAN~\cite{JAN} & 83.5$\pm$0.3& 88.5$\pm$0.2 & 80.1$\pm$0.3& 85.9$\pm$0.4 & 84.5 \\
        DANN~\cite{DANN} & 85.9$\pm$0.4 & 90.5$\pm$0.3 & 88.6$\pm0.4$& 90.4$\pm$0.2& 88.9  \\
        \textbf{DADA} (Ours) & 86.3$\pm$0.3 &91.7$\pm$0.4 & 89.9$\pm$0.3 & \underline{91.3}$\pm$0.3 & \textbf{89.8}\\
        \Xhline{1pt}
        ResNet~\cite{resnet} & 90.5$\pm$0.3 & 94.3$\pm$0.2 & 88.7$\pm$0.4 &82.5$\pm$0.3 & 89.0\\
        SE~\cite{SE} & 90.3$\pm$0.4 & 94.7$\pm$0.4 & 88.5$\pm$0.3 & 85.3$\pm$0.4 & 89.7 \\
        MCD~\cite{MCD_2018} & 91.7$\pm$0.4 &\textbf{95.3}$\pm$0.3 & 89.5$\pm$0.2 & 84.3$\pm$0.2  & 90.2 \\ 
        DANN~\cite{DANN} & 91.5$\pm$0.4 &94.3$\pm$0.4&\underline{90.5}$\pm$0.3&86.3$\pm$0.3 & 90.6\\
        \textbf{DADA} (Ours) &\textbf{92.0}$\pm0.4$& \underline{95.1}$\pm0.3$ & \textbf{91.3}$\pm$0.4& \textbf{93.1}$\pm$0.3&\textbf{92.9}\\
	     \hline
        
        \Xhline{1pt}
    \end{tabular}

\end{table*}

\vspace{0.2cm}
\begin{figure*}[t]
    \begin{minipage}{\hsize}
      \centering
      \subfigure[\scriptsize $\mathcal{A}$-Distance ]
      {\includegraphics[width=0.215\hsize]{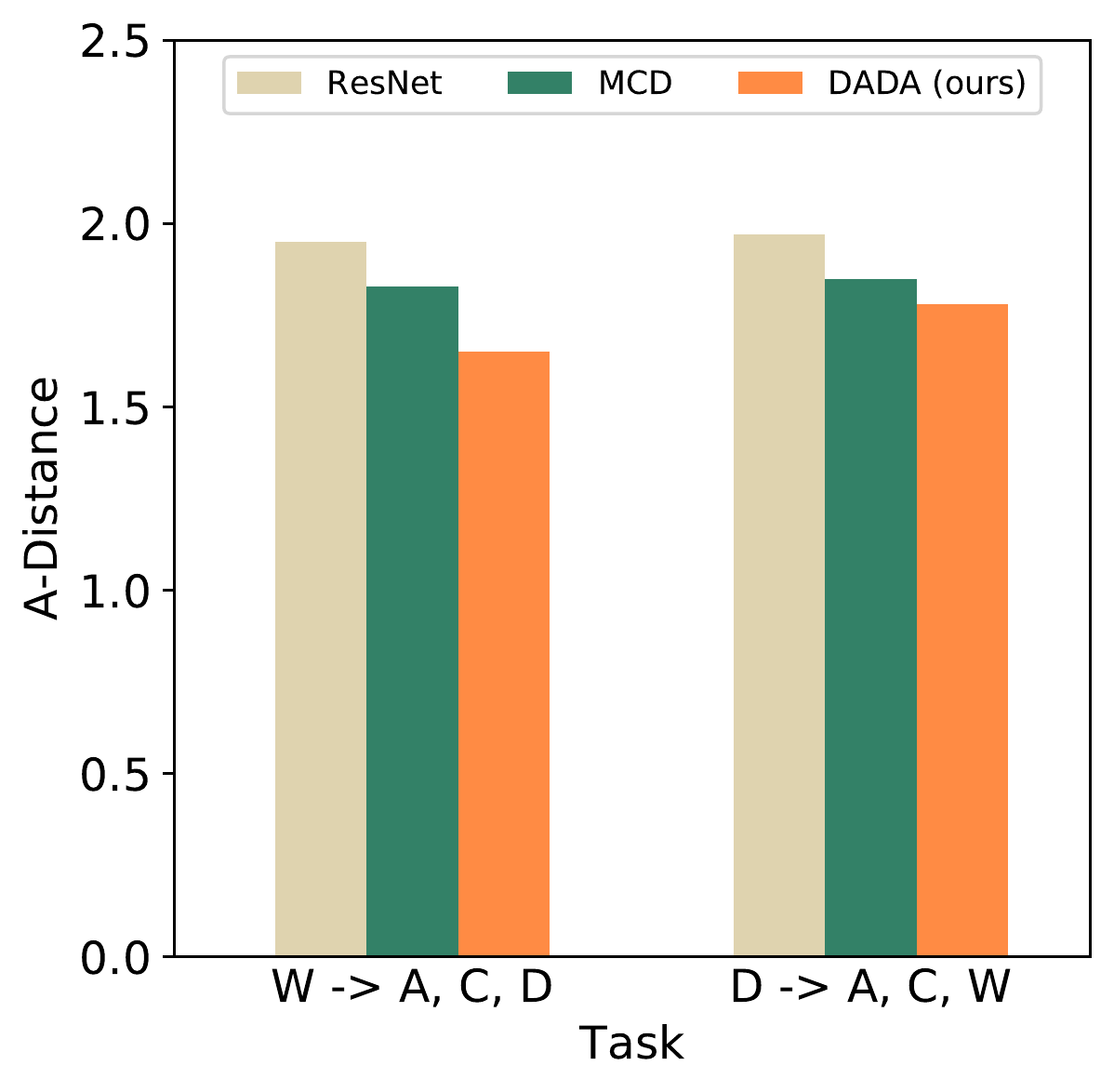}
      \label{a_distance} }
     \centering
      \subfigure[\scriptsize Training loss for C$\rightarrow$A,D,W]
      {\includegraphics[width=0.272\hsize]{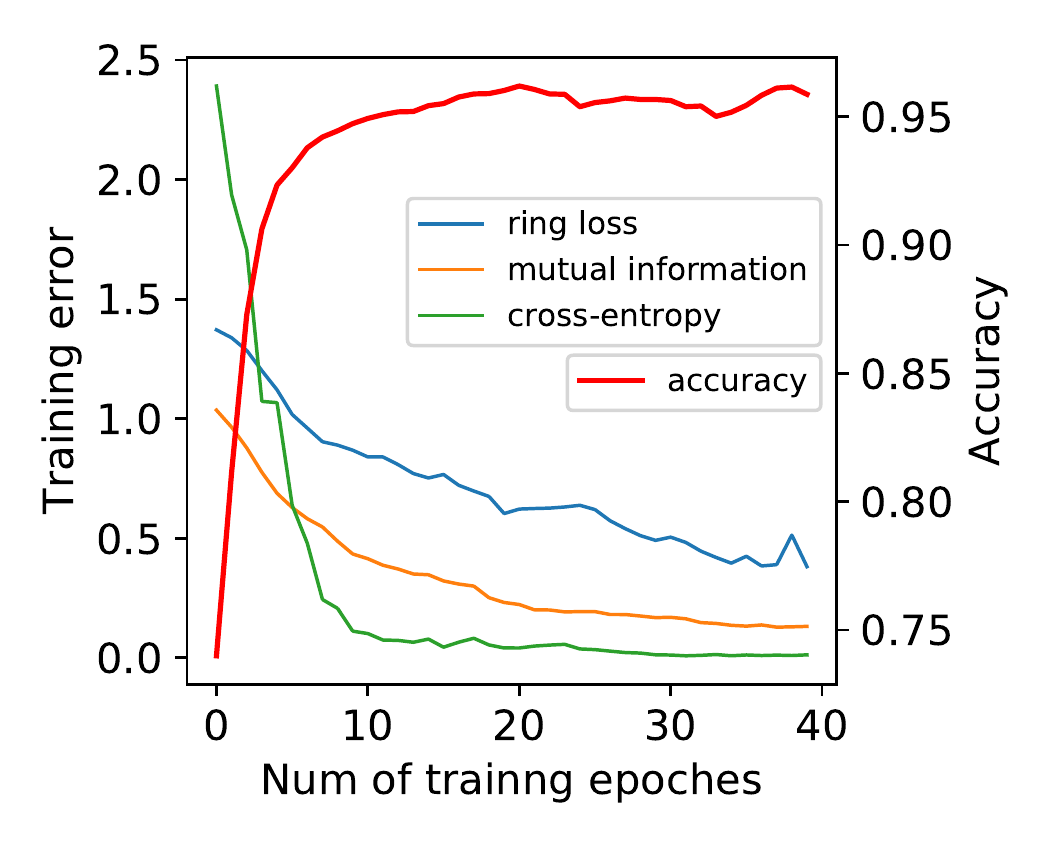}
      \label{office_loss}}
      \centering
      \subfigure[\scriptsize MCD confusion matrix ]
      {\includegraphics[width=0.22\hsize]{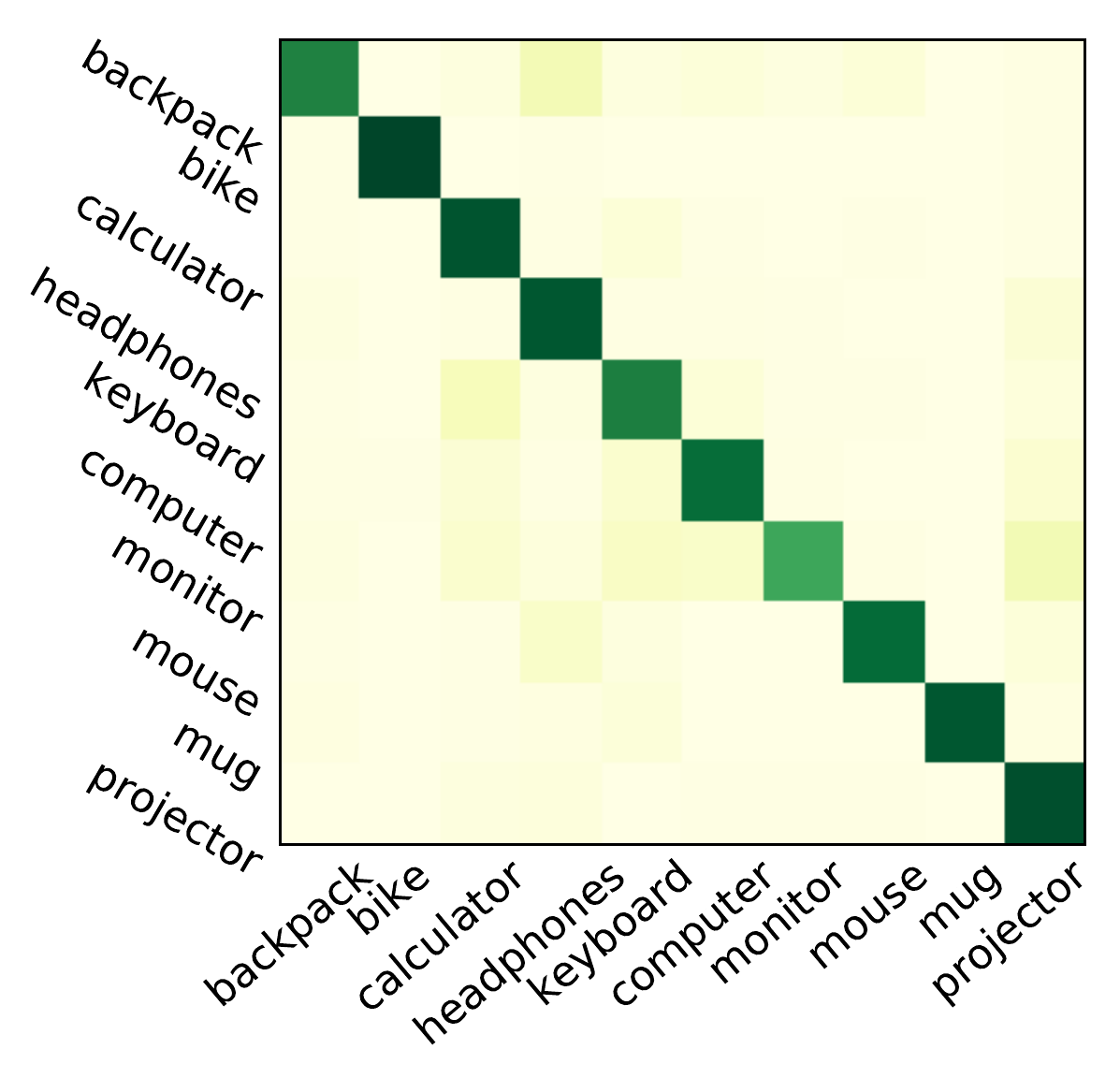}
      \label{office_confusion_mcd}}
      \centering
      \subfigure[\scriptsize DADA confusion matrix ]
      { \includegraphics[width=0.252\hsize]{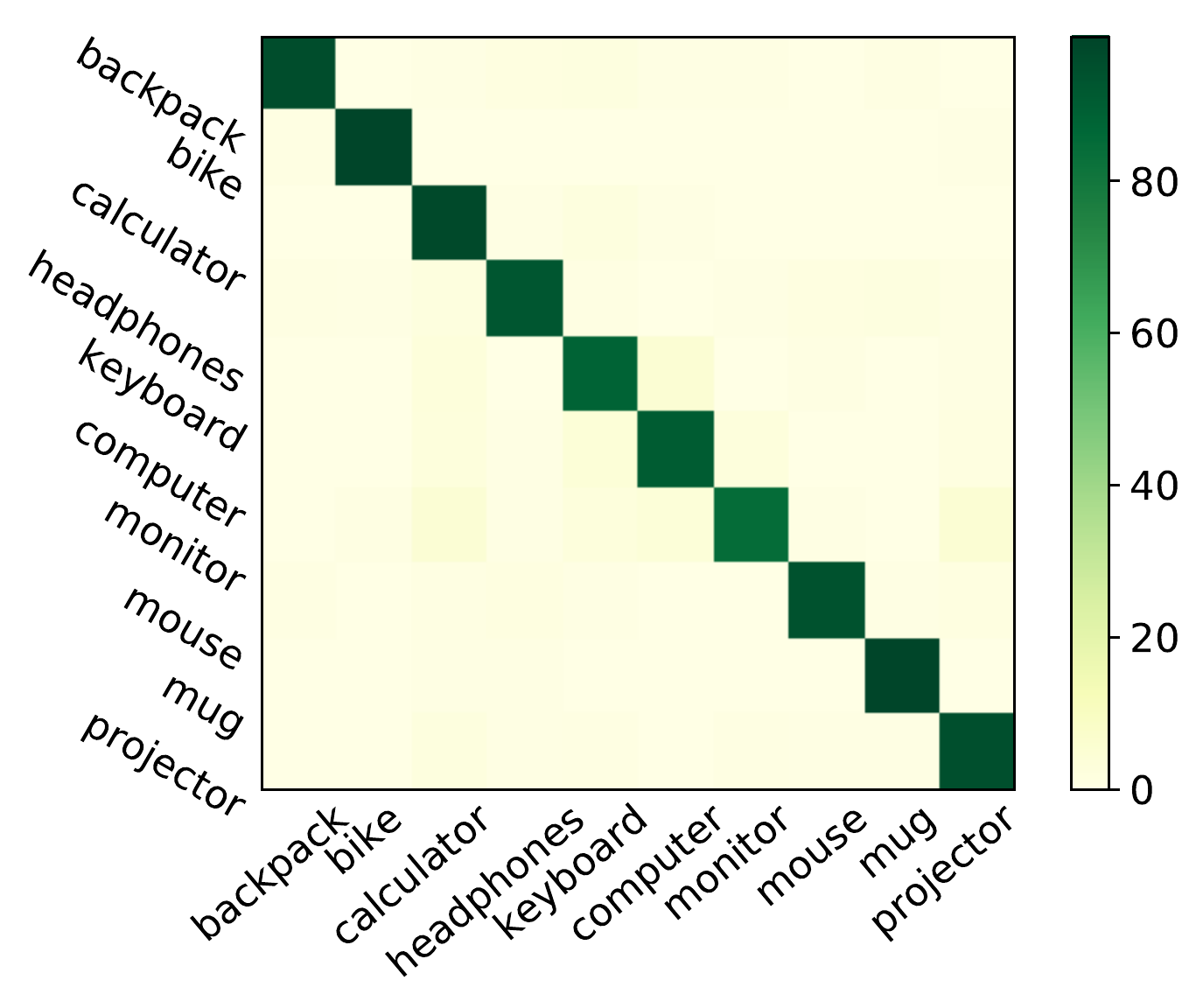}
      \label{office_confusion_dada}}
    \end{minipage}
    \vspace{-0.4cm}
  \caption{Empirical analysis: (a)$\mathcal{A}$-Distance of ResNet, MCD and DADA features on two different tasks; (b) training errors and accuracy on C$\rightarrow$A,D,W task. (c)-(d) confusion matrices of MCD, and DADA models on W$\rightarrow$A,C,D task. }
  \label{fig_office_confusion}
  
\end{figure*}

We compare our model to state-of-the-art baselines: Deep Adaptation Network (\textbf{DAN})~\cite{long2015}, Domain Adversarial Neural Network (\textbf{DANN})~\cite{DANN}, Adversarial Discriminative Domain Adaptation (\textbf{ADDA})~\cite{adda}, Maximum Classifier Discrepancy (\textbf{MCD})~\cite{MCD_2018}, and Unified Feature Disentangler Network (\textbf{UFDN})~\cite{ufdn}. Specifically, DAN applies MMD loss~\cite{gretton2007kernel} to align the source domain with the target domain in reproducing kernel Hilbert space. DANN and ADDA align the source domain with target domain by adversarial loss. MCD is a domain adaptation framework which incorporates two  classifiers. UFDN employs a variational autoencoder~\cite{vae} to disentangle domain-invariant representations. When conducting the baseline experiments, we utilize the code provided by the authors and keep the original experimental settings. 

\textbf{Results and Analysis} The experimental results on the ``Digit-Five'' dataset are shown in Table~\ref{table_digit_five}. From these, we can make the following observations. (1) Model \textbf{\RomanNumeralCaps{4}} achieves 62.3\% average accuracy, significantly outperforming other baselines on most of the domain-agnostic tasks. (2) The results of model \textbf{\RomanNumeralCaps{1}} and \textbf{\RomanNumeralCaps{2}} demonstrate the effectiveness of \textit{class disentanglement} and \textit{domain disentanglement}. Without minimizing the mutual information between disentangled features, UFDN performs poorly on this task. (3) In model \textbf{\RomanNumeralCaps{3}}, the ring loss boost the performance by three percent, demonstrating that feature normalization is essential in domain-agnostic learning.

To dive deeper into the disentangled features, we plot in Figure~\ref{tsne_source}-\ref{tsne_dada} the t-SNE embeddings of the feature representations learned on the {\BV{sv}}$\rightarrow${\BV{mm,mt,up,sy}} task with source-only features, UFDN features, MCD features, and DADA features, respectively. We observe that the features derived by our model are more separated between classes than UFDN and MCD features.

\begin{table*}[t]
\vspace{-0.3cm}
\caption{Accuracy on the DomainNet dataset~\cite{domainnet} dataset with DAL protocol. The table below shows the results based on AlexNet~\cite{alexnet} backbone and the below are the results of ResNet~\cite{resnet} backbone. For both setting, our model outperforms other baselines. } \label{table_lsdac}
\vspace{0.1in}
\centering
\small
{
\begin{tabular}{c c c c c c  c c}
\Xhline{1.0pt}
\multirow{2}{0.9cm}{Models} & 
\multirow{2}{1.4cm}{\scriptsize{\BV{clp$\rightarrow$inf,pnt\\qdr,rel,skt}} } &  
\multirow{2}{1.4cm}{\scriptsize{\BV{inf$\rightarrow$clp,pnt,\\qdr,rel,skt}} } & 
\multirow{2}{1.4cm}{\scriptsize{\BV{pnt$\rightarrow$clp,inf,\\qdr,rel,skt}} }& 
\multirow{2}{1.4cm}{\scriptsize{\BV{qdr$\rightarrow$clp,inf,\\pnt,rel,skt}} }&    
\multirow{2}{1.4cm}{\scriptsize{\BV{rel$\rightarrow$clp,inf,\\pnt,qdr,skt }} } & 
\multirow{2}{1.4cm}{\scriptsize{\BV{skt$\rightarrow$clp,inf,\\pnt,qdr,rel }} } & 
\multirow{2}{0.5cm}{{\BU{Avg}}} \\ 
&&&&&&& \\

 \Xhline{0.7pt} 
AlexNet~\cite{alexnet}& 22.5$\pm$0.4&15.3$\pm$0.2&21.2$\pm$0.3&6.0$\pm$0.2&17.2$\pm$0.3&21.8$\pm$0.3& 17.3\\
DAN~\cite{long2015} & 23.7$\pm$0.3 & 14.9$\pm$0.4 & 22.7$\pm$0.2 & 7.6$\pm$0.3& 19.4$\pm$0.4&\underline{23.4}$\pm$0.5 & 18.6 \\
RTN~\cite{RTN} & 21.4$\pm$0.3 & 14.2$\pm$0.3 & 21.0$\pm$0.4 & 7.7$\pm$0.2 & 17.8$\pm$0.3 &20.8$\pm$0.4  &17.2\\
JAN~\cite{JAN} &21.1$\pm$0.4&16.5$\pm$0.2&21.6$\pm$0.3&9.9$\pm$0.1&15.4$\pm$0.2&22.5$\pm$0.3&17.8\\
DANN~\cite{DANN}&24.1$\pm$0.2 & 15.2$\pm$0.4 & 24.5$\pm$0.3 & 8.2$\pm$0.4 & 18.0$\pm$0.3 &  24.1$\pm$0.4 & 19.1 \\
\textbf{DADA} (Ours) &23.9$\pm$0.4 & 17.9$\pm$0.4 & 25.4$\pm$0.5 & 9.4$\pm$0.2 &20.5$\pm$0.3& \textbf{25.2}$\pm$0.4 & 20.4\\
 \Xhline{1pt} 
 ResNet101~\cite{resnet}& \underline{25.6}$\pm$0.2 & 16.8$\pm$0.3 & 25.8$\pm$0.4 & 9.2$\pm$0.2 & 20.6$\pm$0.5 & 22.3$\pm$0.1 & 20.1  \\
SE~\cite{SE}&21.3$\pm$0.2&8.5$\pm$0.1 &14.5$\pm$0.2&\textbf{13.8}$\pm$0.4 &16.0$\pm$0.4&19.7$\pm$0.2 & 15.6  \\
MCD ~\cite{MCD_2018} & 25.1$\pm$0.3 & \underline{19.1}$\pm$0.4 & \textbf{27.0}$\pm$0.3 & 10.4$\pm$0.3 & 20.2$\pm$0.2 & 22.5$\pm$0.4 & 20.7  \\

\textbf{DADA} (Ours) & \textbf{26.1}$\pm$0.4 &\textbf{20.0}$\pm$0.3 & \underline{26.5}$\pm$0.4 & 12.9$\pm$0.4 & \textbf{20.7}$\pm$0.4 & 22.8$\pm$0.2 &\textbf{21.5}\\

\Xhline{1.0pt}
\end{tabular}
}

\end{table*}

\begin{table}[t]

\caption{One-to-one (o-o) \textit{vs.} one-to-many alignment (o-m). We only show the source domain in the table, the remaining five domains set as the target domain. }
\label{table_one_to_one_vs_one_to_many}
\vspace{0.1in}
\centering
\resizebox{\linewidth}{!}{

{
\begin{tabular}{ccccccccc}
\Xhline{1.0pt}

Source & {\BV{clp}} & {\BV{inf}} & {\BV{pnt}} & {\BV{qdr}} & {\BV{rel}} & {\BV{skt}} & {\BU{Avg}} \\

 \Xhline{0.7pt} 

DAN (\textit{o-o})& 25.2 & 14.9 & 24.1 & 7.8 & 20.4 & 25.2 & 19.6\\
DAN (\textit{o-m})& 23.7 & 14.9 & 22.7&7.6&19.4&23.4&18.6 \\
\Xhline{0.7pt}
JAN (\textit{o-o}) & 24.2 & 18.1 & 23.2 & 7.8 & 15.8 & 23.8 & 18.8 \\
JAN (\textit{o-m})& 21.1 & 16.5 & 21.6 & 9.9 & 15.4 & 22.5 & 17.8 \\

\Xhline{1.0pt}
\end{tabular}

} }

\end{table}

\subsection{Experiments on Office-Caltech10}
\label{exp_office}
\textbf{Office-Caltech10}~\cite{gong2012geodesic} This dataset includes 10 common categories shared by Office-31~\cite{office} and Caltech-256 datasets~\cite{griffin2007caltech}. It contains four domains: \textit{Caltech} (\textbf{C}), which are sampled from Caltech-256 dataset, \textit{Amazon} (\textbf{A}), which contains images collected from \url{amazon.com}, \textit{Webcam} (\textbf{W}) and \textit{DSLR} (\textbf{D}), which are images taken by web camera and DSLR camera under office environment. In our experiments, we take turns to set one domain as the source domain and the rest as the heterogeneous target domain, leading to four DAL tasks. 

In our experiments, we leverage two popular networks, AlexNet~\cite{alexnet} and ResNet~\cite{resnet}, as the backbone of the feature generator $G$. Both the networks are pre-trained on  ImageNet~\cite{ImageNet}. Other components are randomly initialized with normal distribution. In the optimization procedure, we set the learning rate of randomly initialized parameters ten times of the pre-trained parameters. 
The architecture of other components can be seen in Table~\ref{tab:image_arch} (supplementary material). 

In addition to the baselines mentioned in Section~\ref{exp_digit}, we add three baselines: Residual Transfer Network (\textbf{RTN})~\cite{RTN}, Joint Adaptation Network (\textbf{JAN})~\cite{JAN} and Self Ensembling~\cite{SE}. Specifically, RTN employs residual layer~\cite{resnet} for better knowledge transfer, based on DAN~\cite{long2015}. JAN leverages a joint MMD-loss layer to align the features in two consecutive layers. SE applies self-ensembling learning based on a teacher-student model and was the winner of the Visual Domain Adaptation Challenge~\footnote{http://ai.bu.edu/visda-2017/}. We do not apply these methods in digit recognition because the LeNet-based model~\cite{lecun89} is too simple to add a residual or joint training layer. We also omit ADDA and UFDN baselines as these models fail to converge while training on Office-Caltech10 under the domain-agnostic setup. 

\textbf{Results} The experimental results on Office-Caltech10 dataset are shown in Table~\ref{table_office}. For fair comparison, we utilize the same backbone as the baselines and separately show the results. From these results, we make the following observations. (1) Our model achieves \textbf{89.8\%} accuracy with an AlexNet backbone~\cite{alexnet} and \textbf{92.9\%} accuracy with a ResNet backbone, outperforming the corresponding baselines on most shifts. 
(2) The adversarial method (DANN) works better than the feature alignment methods (DAN, RTN, JAN). More interestingly, negative transfer~\cite{pan2010survey} occurs for feature alignment methods. This is somewhat expected, as these models align the entangled features directly, including the class-irrelevant features. (3) From the ResNet results, we observe limited improvements for the baselines from the source-only model, especially for boosting-based SE method. This phenomenon suggests that the boosting procedure works poorly when the target domain is heterogeneously distributed. 

To better analyze the error modes, we plot the confusion matrices for MCD (84.3\% accuracy) and DADA (93.1\% accuracy)  on W$\rightarrow$A,C,D task in Figure~\ref{office_confusion_mcd}-\ref{office_confusion_dada}. The figures illustrate MCD mainly confuses  ``calculator'' \textit{vs.} ``keyboard'',  ``backpack'' \textit{vs.} ``headphones'', and ``monitor'' \textit{vs.} ``projector'', while DADA is able to distinguish them with disentangled features.

\textbf{$\mathcal{A}$-Distance} Ben-David et al.~\yrcite{ben2010theory} suggests $\mathcal{A}$-distance as a measure of domain discrepancy. Following Long et al. ~\yrcite{long2015}, we calculate the approximate $\mathcal{A}$-distance ${\hat d_{\cal A}} = 2\left( {1 - 2\epsilon } \right)$ for W$\rightarrow$A,C,D and D$\rightarrow$A,C,W tasks, where $\epsilon$ is the generalization error of a two-sample classifier (kernel SVM) trained on the binary problem to distinguish input samples between the source and target domains. Figure~\ref{a_distance} displays ${\hat d}_{\cal A}$ for the two tasks with raw ResNet features, MCD features, and DADA features, respectively. We observe that the ${\hat d}_{\cal A}$ for both MCD features and DADA features are smaller than ResNet features, and the ${\hat d}_{\cal A}$ on DADA features is smaller than ${\hat d}_{\cal A}$ on MCD features, which is in consistent with the quantitative results, demonstrating the effectiveness of our disentangled features. 

\textbf{Convergence Analysis} As DADA involves multiple losses and a complex learning procedure including adversarial learning and disentanglement, we analyze the convergence performance for the C$\rightarrow$A,D,W task, as showed in Figure~\ref{office_loss} (lines are smoothed for easier analysis). We plot the cross-entropy loss on the source domain, ring loss defined by Equation~\ref{equ_ring_loss}, mutual information defined by Equation~\ref{equ_mutual_information}, and the accuracy in the figure. Figure~\ref{office_loss} illustrates that the training losses gradually converge and the accuracy become steady after about 20 epochs of training.

\subsection{Experiments on the DomainNet dataset}
\textbf{DomainNet}\footnote{http://ai.bu.edu/M3SDA/}~\cite{domainnet} This dataset contains approximately 0.6 million images distributed among 345 categories. It contains six distinct domains: \textit{Clipart} ({\BV{clp}}), a collection of clipart images; \textit{Infograph} ({\BV{inf}}), infographic images with specific object; \textit{Painting} ({\BV{pnt}}), artistic depictions of object in the form of paintings; \textit{Quickdraw} ({\BV{qdr}}), drawings from the worldwide players of game ``Quick Draw!"\footnote{https://quickdraw.withgoogle.com/data}; \textit{Real} ({\BV{rel}}, photos and real world images; and \textit{Sketch} ({\BV{skt}}), sketches of specific objects. It is very large-scale and includes rich informative vision cues across different domains, providing a good testbed for DAL. Sample images can be seen from Figure~\ref{fig_dataset_overview}. Following Section~\ref{exp_office}, we take turns to set one domain as the source domain and the rest as the heterogeneous target domain, leading to six DAL tasks.

\textbf{Results} The experimental results on DomainNet~\cite{domainnet} are shown in Table~\ref{table_lsdac}. The results shows our model achieves \textbf{21.5\%} accuracy with a ResNet backbone. Note that this dataset contains about 0.6 million images, and so a one percent accuracy improvement is not a trivial achievement. Our model gets comparable results with the best-performing baseline when the source domain is {\BV{pnt}}, or {\BV{qdr}} and outperforms other baselines for the rest of the tasks. From the experimental results, we make two interesting observations. (1) In DAL, the SE model~\cite{SE} performs poorly when the number of categories is large, which is in consistent with results in~\cite{domainnet}. (2) The adversarial alignment method (DANN) performs better than feature alignment methods in DAL, a similar trend to that in Section~\ref{exp_office}.

\textbf{One-to-one \textit{vs.} one-to-many alignment} In the DAL task, the UDA models are performing one-to-many alignment as the target data have no domain labels. However, traditional feature alignment methods such as DAN and JAN are designed for one-to-one alignment. To investigate the effectiveness of domain labels, we design a controlled experiment for DAN and JAN. First, we provide the domain labels and perform one-to-one unsupervised domain adaptation. Then we take away the domain labels and perform one-to-many domain-agnostic learning. The results are shown in Table~\ref{table_one_to_one_vs_one_to_many}. We observe the one-to-one alignment does indeed outperform one-to-many alignment, even though the models in one-to-many alignment have seen more data. These results further demonstrate that DAL is a more challenging task and that traditional feature alignment methods need to be re-thought for this problem.
\section{Conclusion}
In this paper, we first propose a novel domain agnostic learning (DAL) schema and demonstrate the importance of DAL in practical scenarios. Towards tackling DAL task, we have proposed a novel Deep Adversarial Disentangled Autoencoders (DADA) to disentangle \textit{domain-invariant} features in the latent space. 
We have proposed to leveraging \textit{class disentanglement} and mutual information minimizer to enhance the feature disentanglement. Empirically, we demonstrate that the ring-loss-style normalization boosts the performance of DADA in DAL task. An extensive empirical evaluation on DAL benchmarks demonstrate the efficacy of the proposed model against several state-of-the-art domain adaptation algorithms.

\section{Acknowledgement}

We thank Saito Kuniaki, Ben Usman, Ping Hu for their useful discussions and suggestions. We thank anonymous reviewers and area chairs for their useful insight to improve this work. This work was partially supported by NSF and Honda Research Institute.

\bibliography{icml_2019}

\begin{thebibliography}{58}
\providecommand{\natexlab}[1]{#1}
\providecommand{\url}[1]{\texttt{#1}}
\expandafter\ifx\csname urlstyle\endcsname\relax
  \providecommand{\doi}[1]{doi: #1}\else
  \providecommand{\doi}{doi: \begingroup \urlstyle{rm}\Url}\fi

\bibitem[Belghazi et~al.(2018)Belghazi, Baratin, Rajeshwar, Ozair, Bengio,
  Courville, and Hjelm]{mine}
Belghazi, M.~I., Baratin, A., Rajeshwar, S., Ozair, S., Bengio, Y., Courville,
  A., and Hjelm, D.
\newblock Mutual information neural estimation.
\newblock In Dy, J. and Krause, A. (eds.), \emph{Proceedings of the 35th
  International Conference on Machine Learning}, volume~80 of \emph{Proceedings
  of Machine Learning Research}, pp.\  531--540, Stockholmsmässan, Stockholm
  Sweden, 10--15 Jul 2018. PMLR.
\newblock URL \url{http://proceedings.mlr.press/v80/belghazi18a.html}.

\bibitem[Ben-David et~al.(2010)Ben-David, Blitzer, Crammer, Kulesza, Pereira,
  and Vaughan]{ben2010theory}
Ben-David, S., Blitzer, J., Crammer, K., Kulesza, A., Pereira, F., and Vaughan,
  J.~W.
\newblock A theory of learning from different domains.
\newblock \emph{Machine learning}, 79\penalty0 (1-2):\penalty0 151--175, 2010.

\bibitem[Bengio et~al.(2013)Bengio, Courville, and
  Vincent]{bengio2013representation}
Bengio, Y., Courville, A., and Vincent, P.
\newblock Representation learning: A review and new perspectives.
\newblock \emph{IEEE transactions on pattern analysis and machine
  intelligence}, 35\penalty0 (8):\penalty0 1798--1828, 2013.

\bibitem[Cao et~al.(2018)Cao, Katzir, Jiang, Lischinski, Cohen-Or, Tu, and
  Li]{dida}
Cao, J., Katzir, O., Jiang, P., Lischinski, D., Cohen-Or, D., Tu, C., and Li,
  Y.
\newblock Dida: Disentangled synthesis for domain adaptation.
\newblock \emph{arXiv preprint arXiv:1805.08019}, 2018.

\bibitem[Carlucci et~al.(2018{\natexlab{a}})Carlucci, Russo, Tommasi, and
  Caputo]{agnostic_dg_2018}
Carlucci, F.~M., Russo, P., Tommasi, T., and Caputo, B.
\newblock Agnostic domain generalization.
\newblock \emph{CoRR}, abs/1808.01102, 2018{\natexlab{a}}.
\newblock URL \url{http://arxiv.org/abs/1808.01102}.

\bibitem[Carlucci et~al.(2018{\natexlab{b}})Carlucci, Russo, Tommasi, and
  Caputo]{domain_generalization}
Carlucci, F.~M., Russo, P., Tommasi, T., and Caputo, B.
\newblock Agnostic domain generalization.
\newblock \emph{CoRR}, abs/1808.01102, 2018{\natexlab{b}}.
\newblock URL \url{http://arxiv.org/abs/1808.01102}.

\bibitem[Crammer et~al.(2008)Crammer, Kearns, and Wortman]{crammer2008learning}
Crammer, K., Kearns, M., and Wortman, J.
\newblock Learning from multiple sources.
\newblock \emph{Journal of Machine Learning Research}, 9\penalty0
  (Aug):\penalty0 1757--1774, 2008.

\bibitem[Deng et~al.(2009)Deng, Dong, Socher, Li, Li, and Fei-Fei]{ImageNet}
Deng, J., Dong, W., Socher, R., Li, L.-J., Li, K., and Fei-Fei, L.
\newblock Imagenet: A large-scale hierarchical image database.
\newblock In \emph{Computer Vision and Pattern Recognition, 2009. CVPR 2009.
  IEEE Conference on}, pp.\  248--255. IEEE, 2009.

\bibitem[Duan et~al.(2012)Duan, Xu, and Chang]{duan2012exploiting}
Duan, L., Xu, D., and Chang, S.-F.
\newblock Exploiting web images for event recognition in consumer videos: A
  multiple source domain adaptation approach.
\newblock In \emph{Computer Vision and Pattern Recognition (CVPR), 2012 IEEE
  Conference on}, pp.\  1338--1345. IEEE, 2012.

\bibitem[Finn et~al.(2017)Finn, Abbeel, and Levine]{maml}
Finn, C., Abbeel, P., and Levine, S.
\newblock Model-agnostic meta-learning for fast adaptation of deep networks.
\newblock In Precup, D. and Teh, Y.~W. (eds.), \emph{Proceedings of the 34th
  International Conference on Machine Learning}, volume~70 of \emph{Proceedings
  of Machine Learning Research}, pp.\  1126--1135, International Convention
  Centre, Sydney, Australia, 06--11 Aug 2017. PMLR.
\newblock URL \url{http://proceedings.mlr.press/v70/finn17a.html}.

\bibitem[French et~al.(2018)French, Mackiewicz, and Fisher]{SE}
French, G., Mackiewicz, M., and Fisher, M.
\newblock Self-ensembling for visual domain adaptation.
\newblock In \emph{International Conference on Learning Representations}, 2018.
\newblock URL \url{https://openreview.net/forum?id=rkpoTaxA-}.

\bibitem[Ganin \& Lempitsky(2015)Ganin and Lempitsky]{DANN}
Ganin, Y. and Lempitsky, V.
\newblock Unsupervised domain adaptation by backpropagation.
\newblock In Bach, F. and Blei, D. (eds.), \emph{Proceedings of the 32nd
  International Conference on Machine Learning}, volume~37 of \emph{Proceedings
  of Machine Learning Research}, pp.\  1180--1189, Lille, France, 07--09 Jul
  2015. PMLR.
\newblock URL \url{http://proceedings.mlr.press/v37/ganin15.html}.

\bibitem[Ghifary et~al.(2014)Ghifary, Kleijn, and Zhang]{ghifary2014domain}
Ghifary, M., Kleijn, W.~B., and Zhang, M.
\newblock Domain adaptive neural networks for object recognition.
\newblock In \emph{Pacific Rim international conference on artificial
  intelligence}, pp.\  898--904. Springer, 2014.

\bibitem[Gong et~al.(2012)Gong, Shi, Sha, and Grauman]{gong2012geodesic}
Gong, B., Shi, Y., Sha, F., and Grauman, K.
\newblock Geodesic flow kernel for unsupervised domain adaptation.
\newblock In \emph{Computer Vision and Pattern Recognition (CVPR), 2012 IEEE
  Conference on}, pp.\  2066--2073. IEEE, 2012.

\bibitem[Goodfellow et~al.(2014)Goodfellow, Pouget-Abadie, Mirza, Xu,
  Warde-Farley, Ozair, Courville, and Bengio]{gan}
Goodfellow, I., Pouget-Abadie, J., Mirza, M., Xu, B., Warde-Farley, D., Ozair,
  S., Courville, A., and Bengio, Y.
\newblock Generative adversarial nets.
\newblock In \emph{Advances in neural information processing systems}, pp.\
  2672--2680, 2014.

\bibitem[Gretton et~al.(2007)Gretton, Borgwardt, Rasch, Sch{\"o}lkopf, and
  Smola]{gretton2007kernel}
Gretton, A., Borgwardt, K.~M., Rasch, M., Sch{\"o}lkopf, B., and Smola, A.~J.
\newblock A kernel method for the two-sample-problem.
\newblock In \emph{Advances in neural information processing systems}, pp.\
  513--520, 2007.

\bibitem[Griffin et~al.(2007)Griffin, Holub, and Perona]{griffin2007caltech}
Griffin, G., Holub, A., and Perona, P.
\newblock Caltech-256 object category dataset.
\newblock 2007.

\bibitem[He et~al.(2016)He, Zhang, Ren, and Sun]{resnet}
He, K., Zhang, X., Ren, S., and Sun, J.
\newblock Deep residual learning for image recognition.
\newblock In \emph{Proceedings of the IEEE conference on computer vision and
  pattern recognition}, pp.\  770--778, 2016.

\bibitem[Hoffman et~al.(2014)Hoffman, Darrell, and Saenko]{continuous_DA}
Hoffman, J., Darrell, T., and Saenko, K.
\newblock Continuous manifold based adaptation for evolving visual domains.
\newblock In \emph{Computer Vision and Pattern Recognition (CVPR)}, 2014.

\bibitem[Hoffman et~al.(2018)Hoffman, Tzeng, Park, Zhu, Isola, Saenko, Efros,
  and Darrell]{hoffman2017cycada}
Hoffman, J., Tzeng, E., Park, T., Zhu, J.-Y., Isola, P., Saenko, K., Efros, A.,
  and Darrell, T.
\newblock {C}y{CADA}: Cycle-consistent adversarial domain adaptation.
\newblock In Dy, J. and Krause, A. (eds.), \emph{Proceedings of the 35th
  International Conference on Machine Learning}, volume~80 of \emph{Proceedings
  of Machine Learning Research}, pp.\  1989--1998, Stockholmsmässan, Stockholm
  Sweden, 10--15 Jul 2018. PMLR.
\newblock URL \url{http://proceedings.mlr.press/v80/hoffman18a.html}.

\bibitem[Ioffe \& Szegedy(2015)Ioffe and Szegedy]{batch_normalization}
Ioffe, S. and Szegedy, C.
\newblock Batch normalization: Accelerating deep network training by reducing
  internal covariate shift.
\newblock In \emph{International conference on machine learning}, pp.\
  448--456, 2015.

\bibitem[Kiefer et~al.(1952)Kiefer, Wolfowitz, et~al.]{SGD}
Kiefer, J., Wolfowitz, J., et~al.
\newblock Stochastic estimation of the maximum of a regression function.
\newblock \emph{The Annals of Mathematical Statistics}, 23\penalty0
  (3):\penalty0 462--466, 1952.

\bibitem[Kim et~al.(2017)Kim, Cha, Kim, Lee, and Kim]{kim2017learning}
Kim, T., Cha, M., Kim, H., Lee, J.~K., and Kim, J.
\newblock Learning to discover cross-domain relations with generative
  adversarial networks.
\newblock In Precup, D. and Teh, Y.~W. (eds.), \emph{Proceedings of the 34th
  International Conference on Machine Learning}, volume~70 of \emph{Proceedings
  of Machine Learning Research}, pp.\  1857--1865, International Convention
  Centre, Sydney, Australia, 06--11 Aug 2017. PMLR.
\newblock URL \url{http://proceedings.mlr.press/v70/kim17a.html}.

\bibitem[Kingma \& Ba(2014)Kingma and Ba]{Adam}
Kingma, D.~P. and Ba, J.
\newblock Adam: A method for stochastic optimization.
\newblock \emph{arXiv preprint arXiv:1412.6980}, 2014.

\bibitem[Kingma \& Welling(2013)Kingma and Welling]{vae}
Kingma, D.~P. and Welling, M.
\newblock Auto-encoding variational bayes.
\newblock \emph{arXiv preprint arXiv:1312.6114}, 2013.

\bibitem[Kingma et~al.(2014)Kingma, Mohamed, Rezende, and
  Welling]{kingma2014semi}
Kingma, D.~P., Mohamed, S., Rezende, D.~J., and Welling, M.
\newblock Semi-supervised learning with deep generative models.
\newblock In \emph{Advances in neural information processing systems}, pp.\
  3581--3589, 2014.

\bibitem[Krizhevsky et~al.(2012)Krizhevsky, Sutskever, and Hinton]{alexnet}
Krizhevsky, A., Sutskever, I., and Hinton, G.~E.
\newblock Imagenet classification with deep convolutional neural networks.
\newblock In \emph{Advances in neural information processing systems}, pp.\
  1097--1105, 2012.

\bibitem[LeCun et~al.(1989)LeCun, Boser, Denker, Henderson, Howard, Hubbard,
  and Jackel]{lecun89}
LeCun, Y., Boser, B., Denker, J., Henderson, D., Howard, R., Hubbard, W., and
  Jackel, L.
\newblock Backpropagation applied to handwritten zip code recognition.
\newblock \emph{Neural Computation}, 1989.

\bibitem[LeCun et~al.(1998)LeCun, Bottou, Bengio, and
  Haffner]{lecun1998gradient}
LeCun, Y., Bottou, L., Bengio, Y., and Haffner, P.
\newblock Gradient-based learning applied to document recognition.
\newblock \emph{Proceedings of the IEEE}, 86\penalty0 (11):\penalty0
  2278--2324, 1998.

\bibitem[Lee et~al.(2018)Lee, Tseng, Huang, Singh, and Yang]{drit}
Lee, H.-Y., Tseng, H.-Y., Huang, J.-B., Singh, M., and Yang, M.-H.
\newblock Diverse image-to-image translation via disentangled representations.
\newblock In Ferrari, V., Hebert, M., Sminchisescu, C., and Weiss, Y. (eds.),
  \emph{Computer Vision -- ECCV 2018}, pp.\  36--52, Cham, 2018. Springer
  International Publishing.
\newblock ISBN 978-3-030-01246-5.

\bibitem[Li et~al.(2018{\natexlab{a}})Li, Jialin~Pan, Wang, and
  Kot]{li2018domain}
Li, H., Jialin~Pan, S., Wang, S., and Kot, A.~C.
\newblock Domain generalization with adversarial feature learning.
\newblock In \emph{Proceedings of the IEEE Conference on Computer Vision and
  Pattern Recognition}, pp.\  5400--5409, 2018{\natexlab{a}}.

\bibitem[Li et~al.(2018{\natexlab{b}})Li, Gong, Tian, Liu, and
  Tao]{li2018domain_}
Li, Y., Gong, M., Tian, X., Liu, T., and Tao, D.
\newblock Domain generalization via conditional invariant representation,
  2018{\natexlab{b}}.

\bibitem[Liu et~al.(2018{\natexlab{a}})Liu, Liu, Yeh, and Wang]{ufdn}
Liu, A.~H., Liu, Y., Yeh, Y., and Wang, Y.~F.
\newblock A unified feature disentangler for multi-domain image translation and
  manipulation.
\newblock \emph{CoRR}, abs/1809.01361, 2018{\natexlab{a}}.
\newblock URL \url{http://arxiv.org/abs/1809.01361}.

\bibitem[Liu \& Tuzel(2016)Liu and Tuzel]{cogan}
Liu, M.-Y. and Tuzel, O.
\newblock Coupled generative adversarial networks.
\newblock In \emph{Advances in neural information processing systems}, pp.\
  469--477, 2016.

\bibitem[Liu et~al.(2018{\natexlab{b}})Liu, Yeh, Fu, Wang, Chiu, and
  Wang]{detachandattach}
Liu, Y.-C., Yeh, Y.-Y., Fu, T.-C., Wang, S.-D., Chiu, W.-C., and Wang, Y.-C.~F.
\newblock Detach and adapt: Learning cross-domain disentangled deep
  representation.
\newblock In \emph{Proceedings of the IEEE Conference on Computer Vision and
  Pattern Recognition (CVPR)}, 2018{\natexlab{b}}.

\bibitem[Long et~al.(2015)Long, Cao, Wang, and Jordan]{long2015}
Long, M., Cao, Y., Wang, J., and Jordan, M.
\newblock Learning transferable features with deep adaptation networks.
\newblock In Bach, F. and Blei, D. (eds.), \emph{Proceedings of the 32nd
  International Conference on Machine Learning}, volume~37 of \emph{Proceedings
  of Machine Learning Research}, pp.\  97--105, Lille, France, 07--09 Jul 2015.
  PMLR.
\newblock URL \url{http://proceedings.mlr.press/v37/long15.html}.

\bibitem[Long et~al.(2016)Long, Zhu, Wang, and Jordan]{RTN}
Long, M., Zhu, H., Wang, J., and Jordan, M.~I.
\newblock Unsupervised domain adaptation with residual transfer networks.
\newblock In \emph{Advances in Neural Information Processing Systems}, pp.\
  136--144, 2016.

\bibitem[Long et~al.(2017)Long, Zhu, Wang, and Jordan]{JAN}
Long, M., Zhu, H., Wang, J., and Jordan, M.~I.
\newblock Deep transfer learning with joint adaptation networks.
\newblock In \emph{Proceedings of the 34th International Conference on Machine
  Learning, {ICML} 2017, Sydney, NSW, Australia, 6-11 August 2017}, pp.\
  2208--2217, 2017.
\newblock URL \url{http://proceedings.mlr.press/v70/long17a.html}.

\bibitem[Makhzani et~al.(2016)Makhzani, Shlens, Jaitly, Goodfellow, and
  Frey]{makhzani2015adversarial}
Makhzani, A., Shlens, J., Jaitly, N., Goodfellow, I., and Frey, B.
\newblock Adversarial autoencoders.
\newblock \emph{ICLR workshop}, 2016.

\bibitem[Mansour et~al.(2009)Mansour, Mohri, Rostamizadeh, and
  R]{Mansour_nips2018}
Mansour, Y., Mohri, M., Rostamizadeh, A., and R, A.
\newblock Domain adaptation with multiple sources.
\newblock In Koller, D., Schuurmans, D., Bengio, Y., and Bottou, L. (eds.),
  \emph{Advances in Neural Information Processing Systems 21}, pp.\
  1041--1048. Curran Associates, Inc., 2009.

\bibitem[Mathieu et~al.(2016)Mathieu, Zhao, Zhao, Ramesh, Sprechmann, and
  LeCun]{mathieu2016disentangling}
Mathieu, M.~F., Zhao, J.~J., Zhao, J., Ramesh, A., Sprechmann, P., and LeCun,
  Y.
\newblock Disentangling factors of variation in deep representation using
  adversarial training.
\newblock In \emph{Advances in Neural Information Processing Systems}, pp.\
  5040--5048, 2016.

\bibitem[Odena et~al.(2017)Odena, Olah, and Shlens]{cisac_gan}
Odena, A., Olah, C., and Shlens, J.
\newblock Conditional image synthesis with auxiliary classifier {GAN}s.
\newblock In Precup, D. and Teh, Y.~W. (eds.), \emph{Proceedings of the 34th
  International Conference on Machine Learning}, volume~70 of \emph{Proceedings
  of Machine Learning Research}, pp.\  2642--2651, International Convention
  Centre, Sydney, Australia, 06--11 Aug 2017. PMLR.
\newblock URL \url{http://proceedings.mlr.press/v70/odena17a.html}.

\bibitem[Pan \& Yang(2010)Pan and Yang]{pan2010survey}
Pan, S.~J. and Yang, Q.
\newblock A survey on transfer learning.
\newblock \emph{IEEE Transactions on knowledge and data engineering},
  22\penalty0 (10):\penalty0 1345--1359, 2010.

\bibitem[Peng \& Saenko(2018)Peng and Saenko]{peng2017synthetic}
Peng, X. and Saenko, K.
\newblock Synthetic to real adaptation with generative correlation alignment
  networks.
\newblock In \emph{2018 {IEEE} Winter Conference on Applications of Computer
  Vision, {WACV} 2018, Lake Tahoe, NV, USA, March 12-15, 2018}, pp.\
  1982--1991, 2018.
\newblock \doi{10.1109/WACV.2018.00219}.
\newblock URL \url{https://doi.org/10.1109/WACV.2018.00219}.

\bibitem[Peng et~al.(2018)Peng, Bai, Xia, Huang, Saenko, and Wang]{domainnet}
Peng, X., Bai, Q., Xia, X., Huang, Z., Saenko, K., and Wang, B.
\newblock Moment matching for multi-source domain adaptation.
\newblock \emph{arXiv preprint arXiv:1812.01754}, 2018.

\bibitem[Quionero-Candela et~al.(2009)Quionero-Candela, Sugiyama, Schwaighofer,
  and Lawrence]{datashift_book2009}
Quionero-Candela, J., Sugiyama, M., Schwaighofer, A., and Lawrence, N.~D.
\newblock \emph{Dataset Shift in Machine Learning}.
\newblock The MIT Press, 2009.
\newblock ISBN 0262170051, 9780262170055.

\bibitem[Rezende et~al.(2014)Rezende, Mohamed, and
  Wierstra]{stochastic_icml_2014}
Rezende, D.~J., Mohamed, S., and Wierstra, D.
\newblock Stochastic backpropagation and approximate inference in deep
  generative models.
\newblock In Xing, E.~P. and Jebara, T. (eds.), \emph{Proceedings of the 31st
  International Conference on Machine Learning}, volume~32 of \emph{Proceedings
  of Machine Learning Research}, pp.\  1278--1286, Bejing, China, 22--24 Jun
  2014. PMLR.
\newblock URL \url{http://proceedings.mlr.press/v32/rezende14.html}.

\bibitem[Romijnders et~al.(2018)Romijnders, Meletis, and
  Dubbelman]{domain_agnostic_norm18}
Romijnders, R., Meletis, P., and Dubbelman, G.
\newblock A domain agnostic normalization layer for unsupervised adversarial
  domain adaptation.
\newblock \emph{CoRR}, abs/1809.05298, 2018.
\newblock URL \url{http://arxiv.org/abs/1809.05298}.

\bibitem[Saenko et~al.(2010)Saenko, Kulis, Fritz, and Darrell]{office}
Saenko, K., Kulis, B., Fritz, M., and Darrell, T.
\newblock Adapting visual category models to new domains.
\newblock In \emph{European conference on computer vision}, pp.\  213--226.
  Springer, 2010.

\bibitem[Saito et~al.(2018)Saito, Watanabe, Ushiku, and Harada]{MCD_2018}
Saito, K., Watanabe, K., Ushiku, Y., and Harada, T.
\newblock Maximum classifier discrepancy for unsupervised domain adaptation.
\newblock In \emph{The IEEE Conference on Computer Vision and Pattern
  Recognition (CVPR)}, June 2018.

\bibitem[Sun \& Saenko(2016)Sun and Saenko]{SunS16a}
Sun, B. and Saenko, K.
\newblock Deep {CORAL:} correlation alignment for deep domain adaptation.
\newblock \emph{CoRR}, abs/1607.01719, 2016.
\newblock URL \url{http://arxiv.org/abs/1607.01719}.

\bibitem[Tzeng et~al.(2014)Tzeng, Hoffman, Zhang, Saenko, and Darrell]{ddc}
Tzeng, E., Hoffman, J., Zhang, N., Saenko, K., and Darrell, T.
\newblock Deep domain confusion: Maximizing for domain invariance.
\newblock \emph{arXiv preprint arXiv:1412.3474}, 2014.

\bibitem[Tzeng et~al.(2017)Tzeng, Hoffman, Saenko, and Darrell]{adda}
Tzeng, E., Hoffman, J., Saenko, K., and Darrell, T.
\newblock Adversarial discriminative domain adaptation.
\newblock In \emph{Computer Vision and Pattern Recognition (CVPR)}, volume~1,
  pp.\ ~4, 2017.

\bibitem[Xu et~al.(2018)Xu, Chen, Zuo, Yan, and Lin]{xu2018deep}
Xu, R., Chen, Z., Zuo, W., Yan, J., and Lin, L.
\newblock Deep cocktail network: Multi-source unsupervised domain adaptation
  with category shift.
\newblock In \emph{Proceedings of the IEEE Conference on Computer Vision and
  Pattern Recognition}, pp.\  3964--3973, 2018.

\bibitem[Yi et~al.(2017)Yi, Zhang, Tan, and Gong]{yi2017dualgan}
Yi, Z., Zhang, H.~R., Tan, P., and Gong, M.
\newblock Dualgan: Unsupervised dual learning for image-to-image translation.
\newblock In \emph{ICCV}, pp.\  2868--2876, 2017.

\bibitem[Zellinger et~al.(2017)Zellinger, Grubinger, Lughofer,
  Natschl{\"{a}}ger, and Saminger{-}Platz]{cmd}
Zellinger, W., Grubinger, T., Lughofer, E., Natschl{\"{a}}ger, T., and
  Saminger{-}Platz, S.
\newblock Central moment discrepancy {(CMD)} for domain-invariant
  representation learning.
\newblock \emph{CoRR}, abs/1702.08811, 2017.
\newblock URL \url{http://arxiv.org/abs/1702.08811}.

\bibitem[Zheng et~al.(2018)Zheng, Pal, and Savvides]{ringloss}
Zheng, Y., Pal, D.~K., and Savvides, M.
\newblock Ring loss: Convex feature normalization for face recognition.
\newblock \emph{arXiv preprint arXiv:1803.00130}, 2018.

\bibitem[Zhu et~al.(2017)Zhu, Park, Isola, and Efros]{CycleGAN2017}
Zhu, J.-Y., Park, T., Isola, P., and Efros, A.~A.
\newblock Unpaired image-to-image translation using cycle-consistent
  adversarial networks.
\newblock In \emph{Computer Vision (ICCV), 2017 IEEE International Conference
  on}, 2017.

\end{thebibliography}
\bibliographystyle{icml2019}

\clearpage

\appendix
\twocolumn[
\icmltitle{Supplementary Materials}]
\section{Model Architecture}
We provide the detailed model architecture (Table~\ref{tab:digit_arch} and Table~\ref{tab:image_arch}) for each component in our model: Generator, Disentangler, Domain Classifier, Classifier and MINE.
\begin{table}[!ht]
    \centering
    \caption{Model Architecture for `Digit-Five`. For each convolution layer, we list the input dimension, output dimension, kernel size, stride, and padding. For the fully-connected layer, we provide the input and output dimensions. For drop-out layers, we provide the probability of an element to be zeroed.} \label{tab:digit_arch}
    \vspace{0.1in}
    \begin{tabular}{c|l}
        \noalign{\hrule height 1pt}
        layer & configuration \\
        \noalign{\hrule height 1pt}
        \multicolumn{2}{c}{Feature Generator} \\
        \noalign{\hrule height 1pt}
        1 & Conv2D (3, 64, 5, 1, 2), BN, ReLU, MaxPool \\
        \hline
        2 & Conv2D (64, 64, 5, 1, 2), BN, ReLU, MaxPool \\
        \hline
        3 & Conv2D (64, 128, 5, 1, 2), BN, ReLU \\
        \noalign{\hrule height 1pt}     
        \multicolumn{2}{c}{Disentangler} \\
        \noalign{\hrule height 1pt}
        1 & FC (8192, 3072), BN, ReLU\\
        \hline
        2 &  DropOut (0.5), FC (3072, 2048), BN, ReLU \\
        \noalign{\hrule height 1pt}     
        \multicolumn{2}{c}{Domain Identifier} \\
        \noalign{\hrule height 1pt}
        1 & FC (2048, 256), LeakyReLU \\
        \hline
        2 & FC (256, 2), LeakyReLU \\
        \noalign{\hrule height 1pt}     
        \multicolumn{2}{c}{Class Identifier} \\
        \noalign{\hrule height 1pt}
        1 & FC (2048, 10), BN, Softmax \\
        \noalign{\hrule height 1pt}     
        \multicolumn{2}{c}{Reconstructor} \\
        \noalign{\hrule height 1pt}
        1 & FC (4096, 8192) \\
        \noalign{\hrule height 1pt}     
        \multicolumn{2}{c}{Mutual Information Estimator} \\
        \noalign{\hrule height 1pt}
        fc1\_x & FC (2048, 512) \\
        \hline
        fc1\_y & FC (2048, 512), LeakyReLU \\
        \hline
        2 & FC (512,1)\\
        \noalign{\hrule height 1pt}
    \end{tabular}
\end{table}

\section{Details of datasets}
We provide the detailed information of datasets (Table~\ref{tab:dataset_details}). For Digit-Five and the DomainNet dataset, we provide the train/test split for each domain and for Office-Caltech10, we provide the number of images in each domain.
\begin{table}[!ht]
\caption{Detailed information for datasets} \label{tab:dataset_details}
\vspace{0.1in}
\centering

\resizebox{\linewidth}{!}
{
\begin{tabular}{c|ccccccc}

\Xhline{1pt}

\multicolumn{8}{c}{Digit-Five} \\

\Xhline{1pt}

 Splits& {\BV{mnist}} & {\BV{mnist_m}} & {\BV{svhn}} & {\BV{syn}} & {\BV{usps}} & & Total \\
 \hline
 
 Train & 55,000 & 55,000 & 25,000 & 25,000 & 7,348 & & 167,348\\
 Test &  10,000 & 10,000 & 14,549 & 9,000 & 1,860 & & 37,309 \\
 \Xhline{1pt}
 
\multicolumn{8}{c}{Office-Caltech10} \\

\Xhline{1pt}

 Splits &  &{\BV{amazon}} & {\BV{caltech}} & {\BV{dslr}} & {\BV{webcam}}&  & Total \\
\hline
 Total & & 958 & 1,123 & 157 & 295 &  & 2,533 \\
 \Xhline{1pt}
\multicolumn{8}{c}{DomainNet}\\
\Xhline{1pt}
Splits & {\BV{clp}} & {\BV{inf}} & {\BV{pnt}} & {\BV{qdr}} & {\BV{rel}}& {\BV{skt}}& Total \\
 \hline
Train & 34,019 & 37,087 & 52,867 & 120,750 & 122,563 & 49,115 & 416,401 \\
Test & 14,818 & 16,114 & 22,892 & 51,750 & 52,764 & 21,271 & 179,609 \\
\hline
\end{tabular}
}
\end{table}

\begin{table}[!t]
    \centering
    \caption{Model Architecture for `Office-Caltech10` and `DomainNet`. For each convolution layer, we list the input dimension, output dimension, kernel size, stride, and padding. For the fully-connected layer, we provide the input and output dimensions. For drop-out layers, we provide the probability of an element to be zeroed.}\label{tab:image_arch}
    \vspace{0.1in}
    \resizebox{\linewidth}{!}{
    \begin{tabular}{c|l}
        \noalign{\hrule height 1pt}
        layer & configuration \\
        \noalign{\hrule height 1pt}
        \multicolumn{2}{c}{Feature Generator: ResNet101 or AlexNet} \\
        \noalign{\hrule height 1pt}
        \multicolumn{2}{c}{Disentangler} \\
        \noalign{\hrule height 1pt}
        1 & Dropout(0.5), FC (2048, 2048), BN, ReLU \\
        \hline
        2 & Dropout(0.5), FC (2048, 2048), BN, ReLU \\
        \noalign{\hrule height 1pt}     
        \multicolumn{2}{c}{Domain Identifier} \\
        \noalign{\hrule height 1pt}
        1 & FC (2048, 256), LeakyReLU \\
        \hline
        2 & FC (256, 2), LeakyReLU \\
        \noalign{\hrule height 1pt}     
        \multicolumn{2}{c}{Class Identifier} \\
        \noalign{\hrule height 1pt}
        1 & FC (2048, 10), BN, Softmax \\
        \noalign{\hrule height 1pt}     
        \multicolumn{2}{c}{Reconstructor} \\
        \noalign{\hrule height 1pt}
        1 & FC (4096, 2048) \\
        \noalign{\hrule height 1pt}     
        \multicolumn{2}{c}{Mutual Information Estimator} \\
        \noalign{\hrule height 1pt}
        fc1\_x & FC (2048, 512) \\
        \hline
        fc1\_y & FC (2048, 512), LeakyReLU \\
        \hline
        2 & FC (512,1)\\
        \noalign{\hrule height 1pt}
    \end{tabular}}
\end{table}

\end{document}